\documentclass{article}

\usepackage[main,final,nonatbib]{neurips_2026}
\usepackage[numbers]{natbib}

\usepackage[utf8]{inputenc} %
\usepackage[T1]{fontenc}    %
\usepackage{hyperref}       %
\usepackage{url}            %
\usepackage{booktabs}       %
\usepackage{amsfonts}       %
\usepackage{nicefrac}       %
\usepackage{microtype}      %
\usepackage[table]{xcolor}  %
\usepackage{amsmath}        %
\usepackage{siunitx}

\usepackage{multirow}      %
\usepackage{graphicx}      %
\usepackage{float}
\usepackage{placeins}
\usepackage{dsfont}
\usepackage[title]{appendix}

\title{WASP: Weakly Aligned Spatiotemporal Pairs for\\ Fetal Brain MRI--Ultrasound Learning}

\author{%
  Francesco Correnti$^{1}$ \quad Gabriele Magrini$^{1}$ \quad Marco Mistretta$^{1}$ \\
  \textbf{Niccolò Biondi$^{2}$ \quad Pietro Pala$^{1}$ \quad Alessandro Ramalli$^{1}$} \\
  \textbf{Simona Fiori$^{3}$ \quad Andrew D. Bagdanov$^{1}$ \quad Matteo Lenge$^{3}$} \\[0.5em]
  $^{1}$University of Florence, Italy \quad $^{2}$University of Trento, Italy \\
  $^{3}$Meyer Children's Hospital IRCCS, Florence, Italy \\[0.3em]
  \texttt{\{francesco.correnti, gabriele.magrini, marco.mistretta,}\\
  \texttt{pietro.pala, alessandro.ramalli, andrew.bagdanov\}@unifi.it} \\
  \texttt{niccolo.biondi@unitn.it} \quad
  \texttt{\{simona.fiori, matteo.lenge\}@meyer.it} \\[0.4em]
  \href{https://github.com/miccunifi/WASP}{\texttt{ github.com/miccunifi/WASP}
  }
}

\newcommand{\minisection}[1]{\vspace{0.0in} \noindent {\bf #1}}

\usepackage{wrapfig}
\usepackage{pifont}
\usepackage{amssymb}
\usepackage{makecell}
\usepackage{threeparttable}

\usepackage{xspace}
\newcommand{\method}{WASP\xspace}
\newif\ifarxiv
\arxivtrue %
\newcommand{\rev}[1]{\textcolor{black}{#1}} %

\definecolor{nored}{RGB}{190, 0, 0}    %
\definecolor{yesgreen}{RGB}{0, 150, 0} %

\newcommand{\cmark}{\textcolor{yesgreen}{\ding{51}}}
\newcommand{\xmark}{\textcolor{nored}{\ding{55}}}

\newcommand{\goodup}[1]{\textsuperscript{\textcolor{yesgreen}{\tiny$\uparrow$#1}}}
\newcommand{\gooddown}[1]{\textsuperscript{\textcolor{yesgreen}{\tiny$\downarrow$#1}}}
\newcommand{\badup}[1]{\textsuperscript{\textcolor{nored}{\tiny$\uparrow$#1}}}
\newcommand{\baddown}[1]{\textsuperscript{\textcolor{nored}{\tiny$\downarrow$#1}}}

\newcommand{\metric}[2]{#1\,#2}
\newcommand{\metricbase}[1]{#1\textsuperscript{\textcolor{black!35}{\tiny --}}}

\begin{document}

\maketitle

\begin{abstract}
Magnetic Resonance Imaging (MRI) is widely regarded as the optimal sensor for fetal brain analysis due to its superior \rev{soft-tissue} contrast and anatomical detail. However, its high cost and operational burden make it invasive and difficult to obtain at scale. Ultrasound (US), in contrast, is cheap, safe, and routinely acquired, and as a result it has produced substantially larger datasets and a growing ecosystem of pretrained models. This asymmetry raises a natural question: \textit{Can we teach a US-only model to understand fetal MRI from only a limited set of examples?} The standard recipe, training a foundation model on subject-to-subject paired MRI--US scans, is not viable since no such paired fetal dataset is publicly available. In this paper we address this gap with Weakly Aligned Spatiotemporal Pairs (\method), a framework that formulates cross-modal correspondence as an entropic Optimal Transport problem driven by clinical metadata --- in particular, Gestational Age (GA) and diagnostic planes --- enabling the fitting of a lightweight \rev{alignment module} that lifts MRI representations into the US latent space, without fine-tuning the backbone. Empirically, \method yields its largest gains when MRI is unseen by the model during pretraining (on USFM, GA estimation error drops from $21.9$ to \rev{$17.4$} days and standard plane classification accuracy climbs from $61.9$\% to \rev{$69.0$\%}), while \rev{providing smaller, backbone-dependent} refinements for backbones pretrained on both modalities (e.g., BioMedParse GA estimation error from $6.5$ to \rev{$6.0$} days and SAM-Med2d plane accuracy from $83.3$\% to $88.1$\%). %

\end{abstract}

\section{Introduction}

Modern medical imaging is inherently multimodal, with each sensor offering a different compromise between accessibility, invasiveness, acquisition cost, and diagnostic resolution~\cite{azad2023foundational}. Modalities are complementary by design: those that are easier to acquire tend to be coarser, while those that yield finer anatomical detail are typically more expensive, slower, and harder to obtain at scale.

This \rev{trade-off} is particularly sharp in fetal brain analysis. Magnetic Resonance Imaging (MRI) is the preferred sensor whenever a \rev{high-quality} view of the developing brain is required, owing to its superior \rev{soft-tissue} contrast and its ability to resolve subtle morphological structures throughout gestation~\cite{gholipour2017normative}. Ultrasound (US), in contrast, suffers from speckle noise, acoustic shadowing, operator-dependent views, and offers limited fine-grained detail~\cite{malinger2020isuog}. Yet US is cheap, safe, real-time, and routinely acquired in standard prenatal screening. This has produced a marked asymmetry in what is publicly available to the research community: large fetal US cohorts and a growing ecosystem of US-pretrained foundation models~\cite{jiao2024usfm,zhang2025fully,radhachandran2026us,ambsdorf2025general}, versus scarce fetal MRI data and comparatively few MRI-specific backbones~\cite{dong2025mri}.

This asymmetry creates a concrete need for methods that bridge this cross-domain gap. The practical question is how to make efficient use of the few fetal MRI samples that do exist on top of the much larger pretrained US ecosystem, rather than training MRI-specific models from scratch. The standard recipe for this kind of cross-modal transfer relies on subject-to-subject paired acquisitions in which the same fetus is imaged with both modalities at the same point in time and an alignment objective is supervised on those pairs~\cite{cheng2023sam,zhao2024biomedparse}. However, in fetal imaging this recipe is not viable: no public dataset offers same-subject same-day MRI and US fetal scans, as clinical workflows rarely produce such pairs in the first place.

We thus introduce \textit{Weakly Aligned Spatiotemporal Pairs (\method)} to establish cross-modal correspondence between fetal MRI and fetal US without the need for same-day same-subject data. The key observation behind \method is that fetal brain development follows a tightly constrained spatiotemporal trajectory: at a given Gestational Age, on a given standard diagnostic plane, the anatomy is morphologically consistent across subjects~\cite{namburete2023normative,gholipour2017normative} (see Appendix~\ref{sec:appendix_biology} for a detailed biological overview). Indeed, \method exploits this regularity by using two pieces of metadata that are already routinely recorded in clinical practice, Gestational Age (time) and standard diagnostic plane (space), to define a soft mapping via entropic Optimal Transport \cite{cuturi2013sinkhorn}. The result is a pairing that is weak by construction, since it never claims subject identity, but at the same time anatomically meaningful since it ties together scans that depict the same developmental stage of the same brain structure.

This soft pairing is enough to learn a lightweight alignment on top of frozen backbones, which in turn enables effective knowledge transfer strategies, including domain adaptation onto previously unseen modalities and \rev{teacher-to-student} distillation across model families and different modalities, all without fine-tuning the underlying backbone. \method can teach \rev{US-only} models how to interpret fetal MRI, and to refine multimodal foundation models that have already been exposed to both modalities.

The main contributions of this work are:
\begin{itemize}
    \item we introduce \method, a framework that turns routinely available fetal scan metadata into a soft \rev{cross-modal} pairing via entropic Optimal Transport, removing the need for subject-to-subject paired MRI and US data;
    \item we show that \method can adapt frozen \rev{US-only} foundation models to fetal MRI without backbone fine-tuning, and that the same recipe further refines multimodal foundation models that have already seen both modalities, with consistent gains across different backbones; and
    \item we show that \method improves transfer from teacher to students through a simple intra-modal distillation step.
\end{itemize}

\begin{figure*}
    \centering
    \includegraphics[width=\textwidth]{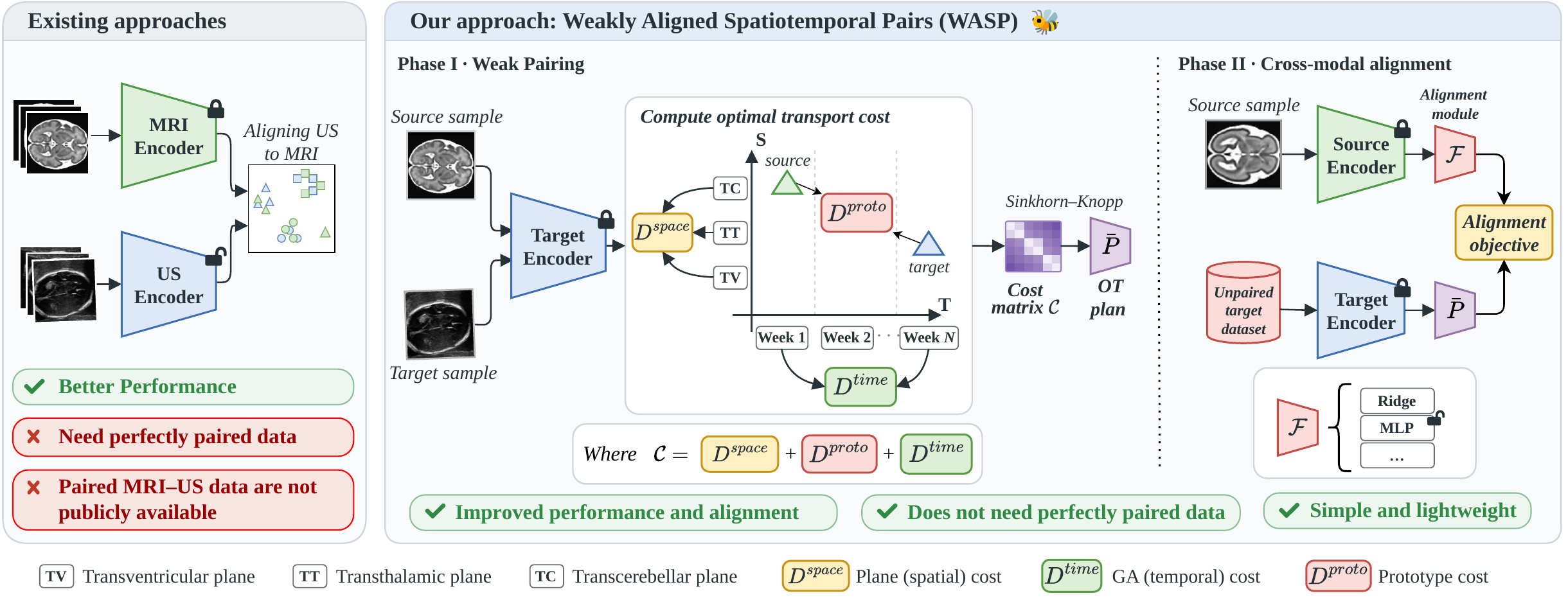}
    \caption{Motivation and overview of \method. \textbf{Left}: Existing MRI–US alignment requires paired fetal scans from the same subject, but such data are not publicly available at scale. \textbf{Right}: \method avoids exact pairing through two phases: \textit{Phase I} builds an entropic Optimal Transport plan ${P}$ from unpaired cohorts using commonly available metadata such as Gestational Age and diagnostic plane. \textit{Phase II} uses ${P}$ to project the target encoder space and train a lightweight alignment module $\mathcal{F}$ on the frozen source encoder's features.}
    \label{fig:teaser}
\end{figure*}

\section{Related Work}

\minisection{Unpaired Multimodal Representation Learning.}
Aligning representations across modalities without paired observations is a \rev{long-standing} problem. One family of approaches synthesizes pseudo-paired data through unpaired image-to-image translation, typically regularized by cycle consistency~\cite{zhu2017unpaired,chen2023deep,silverstein2025translation,jiao2020self}, but this pushes the burden onto a generative model that is prone to hallucination when the source modality is noisy. A second family operates directly in feature space, either by matching distributional moments such as covariance and mean~\cite{sun2017correlation} or by adversarial alignment of latent statistics~\cite{dou2018unsupervised}. Optimal Transport offers a principled alternative that compares full distributions~\cite{courty2016optimal,alvarez2018gromov,demetci2022scot}, and entropic regularization~\cite{cuturi2013sinkhorn,peyre2019computational} makes it tractable at scale.
What unifies these methods is that they treat the two domains as exchangeable point clouds and ignore side information about the samples themselves, which in clinical imaging is precisely what is most reliably available.

\minisection{Prenatal Neuroimaging.}
Fetal neuroimaging is based on two complementary modalities. Ultrasound is fast, portable, and cheap, but is degraded by speckle noise, acoustic shadowing, and \rev{operator-dependent} views~\cite{fiorentino2025uncovering}. MRI yields high \rev{soft-tissue} contrast across the entire gestational window, at the cost of long acquisitions and limited availability~\cite{vahedifard2023review}. This asymmetry is reflected in the modeling landscape: the ultrasound side has seen rapid development of foundation models~\cite{jiao2024usfm,zhang2025fully,radhachandran2026us,ambsdorf2025general,maani2025fetalclip}, while the MRI side is comparatively sparse, with MRI-CORE~\cite{dong2025mri} as a recent generalist backbone. Multimodal medical foundation models such as SAM-Med2d~\cite{cheng2023sam} and BioMedParse~\cite{zhao2024biomedparse} ingest several modalities at once, but treat fetal data as one of many heterogeneous sources with no specific cross-modal alignment mechanism.

\minisection{Bridging US and MRI.} \rev{MRI--US} registration is technically demanding because the two modalities differ in contrast and artifact profile~\cite{fuerst2014automatic,darzi2024review}. Moreover, adult pipelines transfer poorly to fetal data due to motion and rapidly changing anatomy~\cite{marami2017temporal}. Recent fetal-specific work has begun to compare US and MRI phenotypes directly~\cite{wyburd2025cross}, but relies on rare near-same-day acquisitions that are not publicly available at scale.
Several spatiotemporal atlases~\cite{gholipour2017normative,namburete2023normative,calixto2024population,bastiaansen20264d} show that fetal brain growth follows a stereotyped, age-dependent trajectory with well defined population average anatomy for a given gestational week and standard plane. To the best of our knowledge, no prior work uses these population level regularities to define weak correspondences between fully independent US and MRI cohorts, rather than as registration targets for already paired scans. \method fills this gap.

\section{\method: Weakly Aligned Spatiotemporal Pairs}
\label{sec:method}
In this section we describe how \method transforms two unpaired unimodal cohorts into a \rev{weakly paired} multimodal dataset. Let $\mathcal{S} = \{(x^{\mathcal{S}}_i, g^{\mathcal{S}}_i, p^{\mathcal{S}}_i)\}_{i=1}^{N_s}$ and $\mathcal{T} = \{(x^{\mathcal{T}}_j, g^{\mathcal{T}}_j, p^{\mathcal{T}}_j)\}_{j=1}^{N_t}$ denote the source and target cohorts, where $N_{s}$ and $N_{t}$ denote the number of samples (possibly unequal) in the source and target cohorts, respectively. Here, $x \in \mathbb{R}^d$ is a feature embedding extracted from a frozen backbone, $g \in \mathbb{R}$ is the Gestational Age in days, and $p \in \{\text{TV}, \text{TT}, \text{TC}\}$ indicates the standard diagnostic plane—Transventricular (TV), Transthalamic (TT), and Transcerebellar (TC)—as defined by the ISUOG guidelines~\cite{malinger2020isuog}. The roles of source and target are interchangeable, and we describe the construction symmetrically.

The core intuition behind \method is grounded in the regularity of fetal brain development. At a given GA, the brain structures visible on a standard diagnostic plane are anatomically consistent across different subjects: size, morphology, and spatial organization are constrained by the developmental stage~\cite{namburete2023normative, gholipour2017normative} (see Appendix~\ref{sec:appendix_biology} and Fig.~\ref{fig:spatiotemporal_biology}). This inter-subject anatomical regularity makes GA and standard diagnostic plane jointly sufficient to define a meaningful notion of correspondence: two samples acquired at the same age on the same standard diagnostic planes, even from different subjects and different modalities, are expected to depict structurally similar content. Conversely, varying either the GA or the plane breaks this spatiotemporal regularity.

The pair $(g, p)$ is therefore a meaningful spatiotemporal coordinate for cross-subject correspondence, while also being routinely recorded in clinical practice making it a  widely available metadata. \method turns this coordinate into a soft mapping between $\mathcal{S}$ and $\mathcal{T}$ through entropic Optimal Transport, in which a cost matrix encodes temporal distance, \rev{prototype distance}, and a hard spatial consistency term. \rev{Both the Entropic OT, Sinkhorn--Knopp solver and barycentric projection are established tools; the contribution of \method is a cost design that turns population-level developmental regularities into a definition of cross-modal correspondence when subject-level pairs are unavailable.}  Crucially, constructing this weak pairing establishes a common reference space, enabling transfer to unseen modalities, cross-modal knowledge distillation from teacher to student models, and improved performance on downstream clinical tasks.
Fig.~\ref{fig:teaser} shows the full WASP framework, divided in 2 distinct phases. The first phase represents the Weak Pairing, through which the entropic Optimal Transport plan $P$ for the target space is computed. In the second phase, we use the previously obtained $P$ to learn a cross-modal \rev{alignment module} $\mathcal{F}$ from the source to the target space. 

\subsection{Phase I: Weak Pairing}

\minisection{The Spatiotemporal Cost Matrix.}
Given $\mathcal{S}$ and $\mathcal{T}$, we define a composite cost matrix $C \in \mathbb{R}^{N_s \times N_t}$ that penalizes associations between samples that are \textit{temporally} or \textit{spatially dissimilar}: the objective is to find a transport plan $P \in \mathbb{R}_+^{N_s \times N_t}$ that minimizes the total cost induced by $C$.

The \textit{temporal cost} is defined as the absolute GA difference between any \rev{source--target} pair:
\begin{equation}
    D^{\text{time}}_{i,j} = \left| g^{\mathcal{S}}_i - g^{\mathcal{T}}_j \right|
\end{equation}
For the \textit{spatial cost}, we introduce a hard mismatch term:
\begin{equation}
    D^{\text{space}}_{i,j} = \mathds{1}(p^{\mathcal{S}}_i \neq p^{\mathcal{T}}_j)
\end{equation}
where $\mathds{1}(\cdot)$ is the indicator function.

Then, in order to exploit the developmental consistency of fetal brain along with the already \rev{well-structured} target embedding space, we compute a representative prototype for each unique combination of diagnostic plane and gestational window. The core intuition here is that anatomical correspondence holds only within a narrow developmental window; therefore, pooling samples across all gestation would conflate structurally distinct configurations and yield unreliable references.  We therefore partition the target dataset into temporal bins of fixed width (in days) and define the prototype $\mu(g, p)$ as the mean embedding of all target samples within that bin and plane. This aggregation smooths subject-specific variance and grounds the cost in the canonical anatomy of that plane at that developmental stage. Note that prototypes are computed on the target domain, which provides the established reference space into which source representations are aligned. We introduce an additional cost as the Euclidean distance between prototypes\rev{, which we call the \textit{prototype cost}. Note that this cost is not metadata-free: prototypes are indexed by GA bin and plane, so this term encodes spatiotemporal information through the target embedding space}:
\begin{equation}
    \rev{D^{\text{proto}}}_{i,j} = \left\| \mu(g^{\mathcal{S}}_i, p^{\mathcal{S}}_i) - \mu(g^{\mathcal{T}}_j, p^{\mathcal{T}}_j) \right\|_2
\end{equation}

Since $D^{\text{time}}$ and $\rev{D^{\text{proto}}}$ operate on different scales, they are standardized to zero mean and unit variance (denoted $\mathcal{Z}$) before aggregation into the final cost matrix:
\begin{equation}
    \label{eq:cost}
    C_{i,j} = \mathcal{Z}(D^{\text{time}}_{i,j}) + \mathcal{Z}(\rev{D^{\text{proto}}}_{i,j}) + D^{\text{space}}_{i,j}
\end{equation}
To ensure non-negativity, we subtract the global minimum: $\tilde{C} = C - \min(C)$.

\minisection{Soft Optimal Transport Plan.}
We seek an optimal transport plan $P \in \mathbb{R}_+^{N_s \times N_t}$ that minimizes the total alignment cost defined by the cost matrix $\tilde{C}$. Each source sample carries a developmental signature defined by its gestational age and diagnostic plane, and the goal is to match it to target samples that share an analogous developmental and anatomical context. 
The transport plan $P$ encodes the strength of correspondence between source and target, where each entry $P_{i,j}$ quantifies the degree to which source scan $i$ is matched with target scan $j$.

We follow the standard optimal transport formulation~\cite{peyre2019computational}. Since no scan is a priori more informative than another, we assign uniform probability weights to both domains via the marginal vectors $\mathbf{a} = \frac{1}{N_s}\mathbf{1}_{N_s}$ and $\mathbf{b} = \frac{1}{N_t}\mathbf{1}_{N_t}$. These weights act as a fixed budget of correspondence assigned to each scan: every source scan must distribute exactly $1/N_s$ worth of matching across the target set, and every target scan must receive exactly $1/N_t$ worth from the source set. This conservation requirement defines the set of admissible transport plans $\mathcal{U}(\mathbf{a}, \mathbf{b})= \{ \mathcal{P} \in \mathbb{R}_{+}^{N_s \times N_t} \mid \mathcal{P}\mathbf{1}_{N_t} = \mathbf{a}, \; \mathcal{P}^\top\mathbf{1}_{N_s} = \mathbf{b} \},$
which collects all non-negative matrices whose row sums equal $\mathbf{a}$ and whose column sums equal $\mathbf{b}$. Within this set, classical optimal transport seeks the plan that minimizes the linear objective $\langle \mathcal{P}, \tilde{C} \rangle = \sum_{i,j} \mathcal{P}_{i,j} \tilde{C}_{i,j}$. The resulting solution tends to be sparse, producing rigid one-to-one matches between scans. This rigidity is sub-optimal in our setting: biological variability across fetuses means that a single source scan rarely has one perfect counterpart, but rather several plausible matches drawn from neighboring gestational days and the same anatomical plane.

To accommodate this and improve robustness to acquisition noise, we adopt the entropic regularization framework of Cuturi~\cite{cuturi2013sinkhorn}, which augments the standard objective with the Shannon entropy of the plan $H(\mathcal{P})$ .
Maximizing $H(\mathcal{P})$ encourages each source scan to spread its matching budget across several anatomically and temporally compatible target candidates rather than concentrating it on a single counterpart, yielding a smoother routing whose strength is governed by the regularization coefficient $\varepsilon > 0$. Beyond improving robustness, this relaxation admits an efficient GPU-parallel solution via the Sinkhorn--Knopp algorithm (see Sec.~\ref{sec:supp_impl} for implementation details and the $\varepsilon$ hyperparameter search space). The final entropic Optimal Transport plan is therefore obtained as
\begin{equation}
    P = \arg\min_{\mathcal{P} \in \mathcal{U}(\mathbf{a}, \mathbf{b})} \langle \mathcal{P}, \tilde{C} \rangle - \varepsilon H(\mathcal{P}).
\end{equation}

\minisection{Barycentric Projection.}
We then construct for each source embedding a counterpart that lives in the target embedding space, enabling 
direct cross-modal comparison. Concretely, we map each source embedding 
to its counterpart via a convex combination of target embeddings, weighted by the optimal 
transport plan $P$.

Since the rows of the raw plan $P$ do not in general sum to one, we 
row-normalize it to obtain a valid stochastic matrix $\bar{P}:    \bar{P}_{i,j} = {P_{i,j}} \, / \, ({\sum_{k=1}^{N_t} P_{i,k}})$.
Restricting the weights to a convex combination ensures that each target 
embedding computed by the transport plan lies within the convex hull of the target set, preserving its 
geometric structure: if the raw entries of the transport plan $P$ were used directly as weights, the sum of each row would equal $1/N_s$ rather than $1$. Consequently the resulting counterpart embeddings would be scaled down and compressed toward the origin of the vector space. 

The counterpart target embedding matrix is $\tilde{X}^\mathcal{T} = \bar{P} X^\mathcal{T}$, 
while source embeddings remain unchanged $\tilde{X}^\mathcal{S} = X^\mathcal{S}$. 

\subsection{Phase II: Cross-Modal Alignment}
The resulting weakly paired embeddings $(\tilde{X}^\mathcal{S},\, \tilde{X}^\mathcal{T})$ provide the necessary correspondences to learn a cross-modal \rev{alignment module} $\mathcal{F}\colon \mathcal{S} \to \mathcal{T}$ that maps source representations into the target latent space. 
In effect, \method converts a metadata coordinate $(g, p)$ recorded routinely in clinical practice, into a soft correspondence dense enough to train such a mapping, without requiring paired acquisitions at any stage. 
Once $\mathcal{F}$ is learned, source-modality inputs can be projected into the target space, enabling cross-modal transfer and distillation. The specific parametrization of $\mathcal{F}$ and their downstream evaluations are described in Sec.~\ref{sec:exp-prot}.

\section{Experiments}
\label{sec:experiments}
\subsection{Experimental Setup}
\label{sec:exp_setup}
\minisection{Dataset.} 
We consolidate our dataset $\mathcal{D}$ as the union of the ultrasound dataset $\mathcal{D}^{\text{US}}$ and the MRI dataset $\mathcal{D}^{\text{MRI}}$, both consisting of 2D images acquired along the standard diagnostic planes TV, TT, and TC.
We exclusively include neurotypical subjects in both the $\text{US}$ and $\text{MRI}$ cohorts.

The ultrasound dataset contains all the brain images from the \rev{``Multicentre benchmark dataset for comprehensive landmark-based fetal ultrasound biometry''}~\cite{di2025multi}\rev{, which aggregates FETAL\_PLANES\_DB~\cite{burgos2020evaluation}, the HC18 head-circumference dataset~\cite{van2018automated} and an expanded UCLH cohort~\cite{bano2021autofb}; we add the remaining TV and TC images of FETAL\_PLANES\_DB to cover all three planes.} Since US datasets do not provide gestational age directly, $g^{\text{US}}_i$ is estimated from whole-body biometric measurements \rev{with the Hadlock formulas~\cite{hadlock1984estimating}.} We define the ultrasound dataset as $\mathcal{D}^{\text{US}} = \{ (z^{\text{US}}_i, g^{\text{US}}_i, p^{\text{US}}_i) \}_{i=1}^{N_{\text{US}}}$, where each triplet consists of an ultrasound image $z^{\text{US}}_i$, its corresponding Gestational Age $g^{\text{US}}_i$, and the ground truth anatomical plane label $p^{\text{US}}_i$. For additional details regarding the full composition of the dataset and the estimation of gestational age, see Appendix~\ref{sec:supp_data}.

The MRI dataset combines the Fetal Tissue Annotation and segmentation challenge (FeTa)~\cite{zalevskyi2026advances}, the Landmark Fetal Cerebellum (LFC)~\cite{gong2025fetal}, and the Computational Radiology Laboratory (CRL) normative spatiotemporal atlas~\cite{gholipour2017normative} datasets. We leverage already-reconstructed Super-Resolution Reconstruction (SRR) volumes, from which we extract 2D slices along the TV, TT, and TC planes (for further details, see Appendix~\ref{sec:supp_mri_extraction}). We define the MRI dataset analogously to $\mathcal{D}^{\text{US}}$, i.e. $\mathcal{D}^{\text{MRI}} = \{ (z^{\text{MRI}}_j, g^{\text{MRI}}_j, p^{\text{MRI}}_j) \}_{j=1}^{N_{\text{MRI}}}$, where $z^{\text{MRI}}_j$ denotes a 2D slice sampled from an SRR volume. In total, $\mathcal{D}$ comprises $4,056$ images from $1,973$ unique subjects, with  $\mathcal{D}^{\text{US}}$ including $N_{\text{US}} = 3,369$ images from $1,744$ subjects, and  $\mathcal{D}^{\text{MRI}}$ consisting of $N_{\text{MRI}} = 687$ images from $229$ subjects. 

\begin{wrapfigure}[14]{r}{0.50\textwidth}
    \centering
    \vspace{-0.6cm}
    \includegraphics[width=0.5\textwidth]{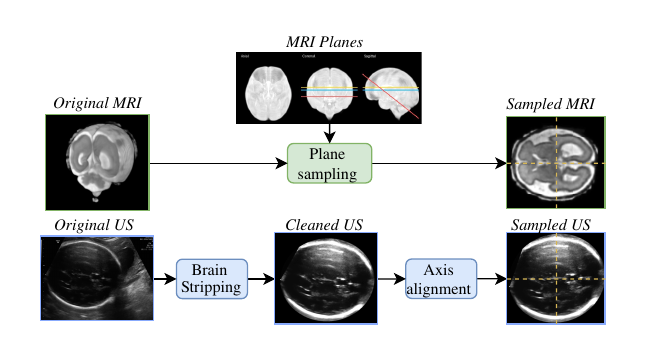}
    \vspace{-0.7cm}
    \caption{US and MRI data preprocessing. While in MRI we can directly get axis-aligned samples, for US images we also need to fix the orientation of the major and minor axes of the fetal brain.
    }\label{fig:sampled}
\end{wrapfigure}
To ensure cross-modal consistency, we apply the data preparation pipeline to all samples in $\mathcal{D}=\{\mathcal{D}^{\text{US}},\mathcal{D}^{\text{MRI}}\}$ summarized in Fig.~\ref{fig:sampled}. Following standard neuroimaging protocols, each sample undergoes brain stripping to remove maternal tissue. This is achieved using a 2-layer CNN decoder head trained on a labeled subsets of the dataset.
The resulting brain-stripped images are oriented such that the major and minor axes of the fetal brain ellipsoid align with the horizontal and vertical axes of the image plane, respectively. Finally, samples from each modality are resampled to their respective isotropic resolutions to ensure spatial consistency.

We partition $\mathcal{D}$ into three disjoint, subject-level subsets: $\mathcal{D}_{\text{pair}}$ is the pairing set used to compute the cross-modal \rev{alignment module} $\mathcal{F}$. 
 $\mathcal{D}_{\text{train}}$ is the training set used to fit the Ridge regression and the SVM classifier.
$\mathcal{D}_{\text{test}}$ is the hold-out evaluation set reserved exclusively for final inference and quantifying the generalization performance of the aligned representations. To tune hyperparameters, we perform \rev{subject-level} $k$-fold cross-validation exclusively on $\mathcal{D}_{\text{train}}$ with $k=3$.

\minisection{Evaluation.}
We define $\theta_{\text{GA}}$ as the parameters of a Ridge regressor for GA prediction, and $\theta_{\text{plane}}$ as the weights of a linear SVM for plane classification. These parameters are fitted on top of the fixed embeddings during the training phase on $\mathcal{D}_{\text{train}}$.
GA Regression is measured by Mean Absolute Error (MAE, in days, lower is better), while Diagnostic Plane Classification is evaluated by accuracy (Cls, \%, higher is better).  To select the optimal hyperparameters, we use a combined metric that balances performance gains across both tasks. For each fold, we compute a normalized score where a positive value indicates improvement over the zero-shot baseline $(\text{MAE}_0, \text{Cls}_0)$: 
$0.5 \cdot \frac{\text{MAE}_0 - \text{MAE}}{\text{MAE}_0} + 0.5 \cdot \frac{\text{Cls} - \text{Cls}_0}{\text{Cls}_0}$. 
We select the configuration that maximizes the median score across the three folds.

\minisection{Models.}
To measure the generalization capabilities of the proposed approach, we evaluated \method using \rev{five} different unimodal and multimodal medical imaging analysis models.
In particular, we chose USFM~\cite{jiao2024usfm} and Echocare~\cite{zhang2025fully} as unimodal models for $\text{US}$ analysis, \rev{MRI-CORE~\cite{dong2025mri} as unimodal model for the $\text{MRI}$ domain, }with SAM-Med2d~\cite{cheng2023sam} and BioMedParse~\cite{zhao2024biomedparse} for multimodal $\text{US}$ and $\text{MRI}$ approaches.

\subsection{Experimental Protocol}
\label{sec:exp-prot}

We compare \method with other spatiotemporal pairing strategies to benchmark its performance. Subsequently, $\mathcal{D}_{\text{train}}$ is used to fit the task-specific parameters $\theta_{\text{GA}}$ and $\theta_{\text{plane}}$. Downstream models are trained on both raw and aligned features to directly quantify the benefit of the alignment. Finally, $\mathcal{D}_{\text{test}}$ serves as the held-out set for all reported evaluations. CCA and Ridge alignments are deterministic; MLP results are reported as mean $\pm$ std over 3 seeds.

\subsubsection{Pairing Strategies}
The pairing strategy determines how source samples $X^{\mathcal{S}}$ are matched to target samples $X^{\mathcal{T}}$ \textit{before} learning the \rev{alignment module} $\mathcal{F}$. We compare \method against two alternative spatiotemporal approaches:

\minisection{Hungarian Spatiotemporal Matching.} A spatiotemporal pairing baseline that uses the same clinical metadata as \method but enforces a strict one-to-one assignment. Within each diagnostic plane, a cost matrix is built from the absolute GA discrepancy between source and target samples, and the Hungarian algorithm computes the $1$-to-$1$ bipartite matching that minimizes the total assignment cost.

\minisection{Greedy Spatiotemporal Matching.} A computationally cheaper baseline that uses the same clinical metadata as Hungarian but \rev{matches each source sample independently. Within each diagnostic plane, every source sample is paired with the target sample having the smallest absolute GA discrepancy. Unlike Hungarian matching, the same target sample may be assigned to multiple source samples.}

\subsubsection{\rev{Alignment Models}}
\rev{We learn $\mathcal{F}$ with CCA, Ridge, or MLP, and include CORAL as a pairing-free baseline.}

\minisection{Correlation Alignment (CORAL).} As a pairing-free baseline, CORAL~\cite{sun2017correlation} aligns the second-order statistics of the raw source and target distributions without requiring any explicit point-to-point matching. The transformation $\mathcal{F}_{\text{coral}}$ is defined by
$
    \rev{W_{\text{coral}} = \Sigma_{\mathcal{S}\mathcal{S}}^{-1/2} \Sigma_{\mathcal{T}\mathcal{T}}^{1/2}}
$
where \rev{$\Sigma_{\mathcal{S}\mathcal{S}}$ and $\Sigma_{\mathcal{T}\mathcal{T}}$} are the regularized covariance matrices of the source and target features, and the aligned representation is obtained via $\mathcal{F}_{\text{coral}}(X^\mathcal{S}) = X^\mathcal{S} W_{\text{coral}}$. 

\minisection{CCA Alignment.} Given paired embeddings $(\tilde{X}^\mathcal{S}, \tilde{X}^\mathcal{T})$, Canonical Correlation Analysis~\cite{hardoon2004canonical} is used to derive a transformation matrix $W$ that maps the source domain directly into the target embedding space. The algorithm computes the whitened cross-covariance \rev{$\mathbf{M} = \Sigma_{\mathcal{S}\mathcal{S}}^{-1/2} \Sigma_{\mathcal{S}\mathcal{T}} \Sigma_{\mathcal{T}\mathcal{T}}^{-1/2}$} and its SVD, \rev{$\mathbf{U}\boldsymbol{\Lambda}\mathbf{V}^\top$}, to find the optimal canonical directions. The final alignment weights are defined as \rev{$W = \Sigma_{\mathcal{S}\mathcal{S}}^{-1/2} \mathbf{U} \mathbf{V}^\top \Sigma_{\mathcal{T}\mathcal{T}}^{1/2}$}. The transformation function \rev{$\mathcal{F}_{\text{cca}}(X^\mathcal{S}) = (X^\mathcal{S} - \bar{x}^{\mathcal{S}}) W + \bar{x}^{\mathcal{T}}$, where $\bar{x}^{\mathcal{S}}$ and $\bar{x}^{\mathcal{T}}$ are the mean source and target embeddings,} aligns source features by first projecting them into the canonical space and then expanding them into the target coordinate system.

\minisection{Ridge Alignment.} An affine transformation $\mathcal{F}_{\text{ridge}}$ is computed by minimizing the ridge-regularized mean squared error between paired embeddings, following the formula $W_{\text{ridge}} = \left((\tilde{X}^\mathcal{S})^\top \tilde{X}^\mathcal{S} + \lambda I\right)^{-1} (\tilde{X}^\mathcal{S})^\top \tilde{X}^\mathcal{T}$, where $\lambda$ is the regularization hyperparameter. The resulting alignment is $\mathcal{F}_{\text{ridge}}(\tilde{X}^\mathcal{S}) = \tilde{X}^\mathcal{S} W_{\text{ridge}}$. For numerical stability, this closed-form solution is computed via SVD of the mean-centered source matrix (see Appendix~\ref{sec:supp_impl} for the $\lambda$ search grid).

\minisection{MLP Alignment.} To capture non-linear cross-modal alignments, we implement an MLP architecture $\mathcal{F}_{\text{mlp}}$ featuring a single hidden layer with GELU activations, optimized via AdamW with weight decay using the MSE loss over the paired embeddings (training hyperparameters and search space in Appendix~\ref{sec:supp_impl}).

\subsection{Experimental Results}

\method can be used in two qualitatively different scenarios. When the source modality is one the backbone has \emph{never seen}, alignment acts as an unlocking mechanism that grafts the unseen modality onto a specialist representation. When the backbone has \emph{already been exposed} to both modalities, alignment instead refines the existing cross-modal geometry.
\rev{Our primary setting is MRI$\to$US ($\mathcal{S}=\mathcal{D}^{\text{MRI}}$, $\mathcal{T}=\mathcal{D}^{\text{US}}$): it matches the motivating asymmetry, where scarce MRI is interpreted through the much larger US ecosystem, and the US cohort provides the larger target space to anchor the transport plan ($961$ vs.\ $480$ pairing images). Tab.~\ref{tab:self_distill_mri2us} reports this direction. Distillation and ablation studies below also use this direction. We then reverse source and target and report US$\to$MRI in Tab.~\ref{tab:self_distill_us2mri}, which adds the MRI-native MRI-CORE~\cite{dong2025mri}.}

\begin{table}
\centering
\caption{%
Cross-modal alignment results in the \rev{\textbf{MRI$\to$US}} direction across four backbones: two unimodal US models (USFM, Echocare) and two multimodal models (SAM-Med2d, BioMedParse). We compare zero-shot transfer, CORAL, and three pairing strategies combined with CCA, Ridge, or MLP alignment. Performance is measured by GA regression error ($\downarrow$) and diagnostic-plane classification accuracy ($\uparrow$). \rev{CORAL, CCA, and Ridge are deterministic; MLP rows report mean $\pm$ std over 3 seeds. Best and second-best results are shown in bold and underlined, respectively.}
}
\label{tab:self_distill_mri2us}
\small
\resizebox{\textwidth}{!}{%
\begin{tabular}{l l  cc  cc cc  cc}
\toprule
\multicolumn{2}{l}{\multirow{2}{*}{\rev{\textbf{MRI$\to$US}}}} & \multicolumn{4}{c}{Unimodal (US)} & \multicolumn{4}{c}{Multimodal}  \\
\cmidrule(lr){3-6} \cmidrule(r){7-10} 
& & \multicolumn{2}{c}{USFM~\cite{jiao2024usfm}} & \multicolumn{2}{c}{Echocare~\cite{zhang2025fully}} & \multicolumn{2}{c}{SAM-Med2d~\cite{cheng2023sam}} & \multicolumn{2}{c}{BioMedParse~\cite{zhao2024biomedparse}} \\
\cmidrule(lr){3-4} \cmidrule(l){5-6} \cmidrule(r){7-8} \cmidrule(lr){9-10}
Pairing & Alignment & GA Reg.\,$\downarrow$ & Plane Cls.\,$\uparrow$ & GA Reg.\,$\downarrow$ & Plane Cls.\,$\uparrow$ & GA Reg.\,$\downarrow$ & Plane Cls.\,$\uparrow$ & GA Reg.\,$\downarrow$ & Plane Cls.\,$\uparrow$ \\
\midrule
Zero-Shot & --- & 21.9 & 61.9 & 10.3 & 81.0 & \textbf{7.7} & 83.3 & 6.5 & \textbf{85.7} \\
\midrule
--- & CORAL~\cite{sun2017correlation} & 22.4 & 54.8 & 29.7 & 69.0 & 41.2 & 50.0 & 24.9 & 40.5 \\
\midrule
\multirow{3}{*}{Greedy}
  & CCA & 25.5 & \underline{66.7} & 16.8 & 76.2 & 21.0 & 83.3 & 14.5 & 78.6 \\
 & Ridge & 25.2 & \underline{66.7} & 10.4 & 83.3 & 10.1 & 85.7 & \underline{6.1} & \textbf{85.7} \\
 & MLP & \rev{24.1{\scriptsize$\pm$0.5}} & \rev{62.7{\scriptsize$\pm$1.4}} & \rev{10.0{\scriptsize$\pm$0.2}} & \rev{81.8{\scriptsize$\pm$1.4}} & \rev{8.1{\scriptsize$\pm$0.9}} & \rev{82.5{\scriptsize$\pm$2.7}} & \rev{8.2{\scriptsize$\pm$0.5}} & \rev{82.5{\scriptsize$\pm$1.4}}\\
\midrule
\multirow{3}{*}{Hungarian}
  & CCA & 28.0 & \rev{59.5} & 17.0 & \rev{73.8} & 38.5 & \rev{59.5} & 26.1 & \rev{61.9}\\
 & Ridge & 21.7 & 61.9 & 11.8 & 83.3 & 15.6 & 85.7 & 11.3 & \textbf{85.7} \\
 & MLP & \rev{20.2{\scriptsize$\pm$0.1}} & \rev{61.9{\scriptsize$\pm$2.4}} & \rev{11.3{\scriptsize$\pm$0.9}} & \rev{80.2{\scriptsize$\pm$1.4}} & \rev{11.1{\scriptsize$\pm$0.7}} & \rev{\textbf{90.5{\scriptsize$\pm$2.4}}} & \rev{12.6{\scriptsize$\pm$1.8}} & \rev{\underline{84.1{\scriptsize$\pm$2.7}}}\\
\midrule
\rowcolor{blue!8}  & CCA & \rev{\underline{17.7}} & \rev{64.3} & \rev{11.9} & \rev{81.0} & \rev{28.3} & \rev{50.0} & \rev{15.2} & \rev{54.8}\\
\rowcolor{blue!8} \textbf{\method (Ours)} & Ridge & \textbf{17.4} & \textbf{69.0} & \textbf{9.1} & \textbf{90.5} & 9.0 & 88.1 & 6.9 & \textbf{85.7} \\
\rowcolor{blue!8}  & MLP & \rev{19.9{\scriptsize$\pm$0.1}} & \rev{65.1{\scriptsize$\pm$1.4}} & \rev{\underline{9.7{\scriptsize$\pm$0.5}}} & \rev{\underline{84.1{\scriptsize$\pm$1.4}}} & \rev{\underline{7.9{\scriptsize$\pm$0.3}}} & \rev{\underline{89.7{\scriptsize$\pm$2.7}}} & \rev{\textbf{6.0{\scriptsize$\pm$0.6}}} & \rev{\underline{84.1{\scriptsize$\pm$1.4}}}\\
\bottomrule
\end{tabular}}
\end{table}

\minisection{Unlocking an unseen modality.}
USFM and Echocare are \rev{pretrained} exclusively on ultrasound; MRI is, for them, a new modality. \method delivers the largest relative gains observed in the paper: \rev{with \method+Ridge,} USFM's GA MAE drops from 21.9 to \rev{17.4} days and plane classification climbs from 61.9\% to \rev{69.0}\%, while Echocare improves to 9.1 days and 90.5\% accuracy. Other approaches, while still moderately improving performance, are less reliable than \method, and in many instances  degrade performance.

\minisection{Refining a multimodal representation.}
Differently from the \rev{unimodal} pretrained models, SAM-Med2d and BioMedParse have already been exposed to both modalities, so the question shifts from unlocking to refinement: can \method improve on a representation that already crosses the \rev{modality} gap? \rev{In the MRI$\to$US direction, \method+MLP achieves the best MAE on BioMedParse ($6.0 \pm 0.6$ vs.\ $6.5$ Zero-Shot) and lifts SAM-Med2d plane classification to $89.7 \pm 2.7$\% (vs.\ $83.3$\% Zero-Shot), with Hungarian+MLP reaching $90.5 \pm 2.4$\%, while \method+Ridge ties Zero-Shot on BioMedParse plane classification at $85.7$\%. With MLP alignment, \method is the only pairing that improves one metric on each multimodal backbone while staying close to Zero-Shot on the other: Greedy+MLP degrades both metrics on both backbones, and Hungarian+MLP substantially increases the GA error.}

\minisection{\rev{Complementary direction: US$\to$MRI.}}
\rev{Reversing the roles maps the native US representations of USFM and Echocare toward the non-native MRI space. MRI-CORE is the mirrored setting, with US as the source modality, although it is already strong on US without any alignment ($12.3$ days, $98.1\%$), on par with the multimodal backbones. Starting from high Zero-Shot performance ($\geq92.8\%$ plane accuracy on every backbone), this direction leaves little headroom: \method+MLP never loses more than $1.4$ days of GA MAE or $0.3$ points of plane accuracy with respect to Zero-Shot, and improves plane classification on four of the five backbones (e.g., SAM-Med2d $97.3\%$ vs.\ $96.2\%$), while most CCA and Ridge variants degrade performance.}

\begin{table}[!ht]
  \centering
  \caption{%
    Cross-modal alignment results in the complementary \textbf{US$\to$MRI} direction across five backbones: two unimodal US models (USFM, Echocare), one unimodal MRI model (MRI-CORE), and two multimodal models (SAM-Med2d, BioMedParse). We report gestational-age regression (GA Reg.; lower is better) and plane classification accuracy (Plane Cls.; higher is better), comparing Zero-Shot performance against different pairing and alignment strategies. \rev{CORAL, CCA, and Ridge are deterministic; MLP rows report mean $\pm$ std over 3 seeds. Best and second-best results within each backbone and task are shown in bold and underlined, respectively.}
    }
  \label{tab:self_distill_us2mri}
  \small
  \resizebox{\textwidth}{!}{%
  \begin{tabular}{l l cc cc cc cc cc}
  \toprule
  \multicolumn{2}{l}{\multirow{2}{*}{\rev{\textbf{US$\to$MRI}}}} & \multicolumn{4}{c}{Unimodal (US)}
    & \multicolumn{2}{c}{Unimodal (MRI)}
    & \multicolumn{4}{c}{Multimodal} \\
  \cmidrule(lr){3-6}
  \cmidrule(lr){7-8}
  \cmidrule(lr){9-12}

  & & \multicolumn{2}{c}{USFM~\cite{jiao2024usfm}}
    & \multicolumn{2}{c}{Echocare~\cite{zhang2025fully}}
    & \multicolumn{2}{c}{MRI-CORE~\cite{dong2025mri}}
    & \multicolumn{2}{c}{SAM-Med2d~\cite{cheng2023sam}}
    & \multicolumn{2}{c}{BioMedParse~\cite{zhao2024biomedparse}} \\

  \cmidrule(lr){3-4}
  \cmidrule(lr){5-6}
  \cmidrule(lr){7-8}
  \cmidrule(lr){9-10}
  \cmidrule(lr){11-12}

  Pairing & Alignment
  & GA Reg.\,$\downarrow$ & Plane Cls.\,$\uparrow$
  & GA Reg.\,$\downarrow$ & Plane Cls.\,$\uparrow$
  & GA Reg.\,$\downarrow$ & Plane Cls.\,$\uparrow$
  & GA Reg.\,$\downarrow$ & Plane Cls.\,$\uparrow$
  & GA Reg.\,$\downarrow$ & Plane Cls.\,$\uparrow$ \\
  \midrule

  Zero-Shot & ---
  & \textbf{16.0} & 92.8
  & \underline{14.9} & \underline{95.6}
  & \textbf{12.3} & \underline{98.1}
  & 15.7 & 96.2
  & \textbf{12.1} & 98.7 \\

  --- & CORAL~\cite{sun2017correlation}
  & 23.7 & 85.0
  & 23.8 & 85.4
  & 16.7 & 96.2
  & 19.0 & 91.1
  & 23.5 & 88.4 \\

  \midrule

  \multirow{3}{*}{Greedy}
  & CCA
  & 23.2 & 89.7
  & 21.5 & 92.2
  & 24.9 & 93.5
  & 26.1 & 93.0
  & 18.6 & 95.8 \\

  & Ridge
  & 22.8 & 90.1
  & 20.6 & 92.6
  & 25.3 & 93.5
  & 22.0 & 96.4
  & 17.9 & 98.5 \\

  & MLP
  & \rev{17.7$\pm$0.2} & \rev{\textbf{93.2$\pm$0.1}}
  & \rev{\textbf{14.7$\pm$0.1}} & \rev{\textbf{96.2$\pm$0.2}}
  & \textbf{12.3$\pm$0.2} & 97.2$\pm$0.1
  & \rev{15.7$\pm$0.1} & \rev{\underline{97.0$\pm$0.2}}
  & \rev{\underline{12.3$\pm$0.2}} & \rev{98.5$\pm$0.2} \\

  \midrule

  \multirow{3}{*}{Hungarian}
  & CCA
  & 24.3 & \rev{71.9}
  & 17.2 & \rev{91.1}
  & 16.1 & 92.0
  & 22.0 & \rev{80.2}
  & 17.4 & \rev{94.3} \\

  & Ridge
  & 20.7 & 80.0
  & 17.2 & 93.9
  & 14.2 & 96.8
  & \textbf{14.2} & 94.5
  & 12.8 & 97.9 \\

  & MLP
  & \rev{17.6$\pm$0.2} & \rev{90.2$\pm$0.1}
  & \rev{15.5$\pm$0.2} & \rev{\textbf{96.2$\pm$0.0}}
  & 13.7$\pm$0.2 & 97.4$\pm$0.1
  & \rev{15.7$\pm$0.1} & \rev{96.2$\pm$0.4}
  & \rev{12.8$\pm$0.2} & \rev{\underline{98.8$\pm$0.1}} \\

  \midrule

  \rowcolor{blue!8}
  & CCA
  & \rev{22.5} & \rev{92.2}
  & \rev{20.1} & \rev{92.2}
  & 24.6 & 92.6
  & \rev{25.4} & \rev{94.9}
  & \rev{18.1} & \rev{97.5} \\

  \rowcolor{blue!8}
  \textbf{\method (Ours)}
  & Ridge
  & 22.0 & 90.1
  & 20.8 & 93.5
  & 25.1 & 93.0
  & 21.9 & 95.1
  & 17.4 & \textbf{98.9} \\

  \rowcolor{blue!8}
  & MLP
  & \rev{\underline{17.4$\pm$0.3}} & \rev{\underline{93.0$\pm$0.4}}
  & \rev{\underline{14.9$\pm$0.3}} & \rev{\textbf{96.2$\pm$0.4}}
  & \underline{12.8$\pm$0.2} & \textbf{98.2$\pm$0.4}
  & \rev{\underline{15.3$\pm$0.3}} & \rev{\textbf{97.3$\pm$0.0}}
  & \rev{12.4$\pm$0.3} & \rev{98.4$\pm$0.4} \\

  \bottomrule
  \end{tabular}}
  \end{table}

\begin{wraptable}[19]{r}{0.5\textwidth}
\centering
\vspace{-0.3cm}
\caption{%
  \rev{\textbf{MRI$\to$US.}} Teacher-student distillation results on the MRI test set.
  $\mathcal{G}_{m \to m}$ maps student embeddings to those of a teacher
  (raw or \method-improved) using a Ridge regressor; the teacher alignment
  $\mathcal{F}$ is \method+MLP. \rev{Raw rows are deterministic; + \method rows report mean $\pm$ std over 3 seeds.}
}
\label{tab:self_distill_teacher}

\footnotesize
\setlength{\tabcolsep}{3pt}        %
\renewcommand{\arraystretch}{0.95} %

\resizebox{\linewidth}{!}{%
\begin{tabular}{c l l r r}
\toprule
Student & Teacher & Setup & GA$\downarrow$ & Cls.\,$\uparrow$ \\
\midrule

\multirow{5}{*}{{USFM~\cite{jiao2024usfm}}}
    & Zero-Shot
    & - & 21.9 & 61.9 \\
    \cmidrule(l){2-5}
  & \multirow{2}{*}{SAM-Med2d~\cite{cheng2023sam}}
    & Raw & 12.4 & \textbf{76.2} \\
  & & \cellcolor{blue!8}+ \method & \cellcolor{blue!8}\rev{12.0{\scriptsize$\pm$0.8}} & \cellcolor{blue!8}\rev{\underline{72.2{\scriptsize$\pm$1.4}}}\\
  \cmidrule(l){2-5}
  & \multirow{2}{*}{BioMedParse~\cite{zhao2024biomedparse}}
    & Raw & \underline{11.9} & 69.0 \\
  & & \cellcolor{blue!8}+ \method & \cellcolor{blue!8}\rev{\textbf{11.1{\scriptsize$\pm$0.2}}} & \cellcolor{blue!8}\rev{71.4{\scriptsize$\pm$0.0}}\\

\midrule

\multirow{5}{*}{{Echocare~\cite{zhang2025fully}}}
    & Zero-Shot
    & - & 10.3 & 81.0 \\
    \cmidrule(l){2-5}
  & \multirow{2}{*}{SAM-Med2d~\cite{cheng2023sam}}
    & Raw & 8.9 & 85.7 \\
  & & \cellcolor{blue!8}+ \method & \cellcolor{blue!8}\rev{\underline{8.8{\scriptsize$\pm$0.1}}} & \cellcolor{blue!8}\rev{\underline{89.7{\scriptsize$\pm$1.4}}}\\
  \cmidrule(l){2-5}
  & \multirow{2}{*}{BioMedParse~\cite{zhao2024biomedparse}}
    & Raw & 9.3 & 88.1 \\
  & & \cellcolor{blue!8}+ \method & \cellcolor{blue!8}\rev{\textbf{8.6{\scriptsize$\pm$0.4}}} & \cellcolor{blue!8}\rev{\textbf{90.5{\scriptsize$\pm$2.4}}}\\
\addlinespace[3pt]
\bottomrule
\end{tabular}}
\end{wraptable}

\minisection{Propagating gains to student backbones.}
The previous experiments treat each backbone independently. We now ask if the gains achieved by aligning a teacher's representation can help distillation between different models. Concretely, we train a transfer function $\mathcal{G}_{m \to m}$ that maps a student's raw embeddings in modality $m \in \{\text{US}, \text{MRI}\}$ to the teacher's improved embedding space in the same modality, parameterized as either a Ridge regressor or an MLP. When the teacher is a large multimodal model (BioMedParse or SAM-Med2d) whose representations are inherently richer, $\mathcal{G}_{m \to m}$ provides a lightweight mechanism to boost the student without modifying its backbone; when the teacher has been previously improved via \method, this boost is amplified\rev{: GA regression improves for every teacher--student pair, and plane classification improves in three of the four}.

\vspace{.2cm}
We chain two stages: (i) the teacher is first improved via \method cross-modal alignment, and (ii) $\mathcal{G}_{m \to m}$ is learned on $\mathcal{D}_{\text{pair}}$ by regressing the student's embeddings onto the teacher's improved representations. At inference time, student features $X^{m}_{\text{test}}$ are passed through $\mathcal{G}_{m \to m}$ before the downstream evaluation. \rev{Tab.~\ref{tab:self_distill_teacher} reports the primary setting: teachers are improved with \method+MLP in MRI$\to$US, and $m=\text{MRI}$.}

\subsection{Ablations}
\label{sec:ablation}

\rev{We ablate the cost terms and the pairing strategy in the primary MRI$\to$US direction; the MLP hyperparameter sensitivity and the MRI-CORE pairing ablation in both directions are in Appendices~\ref{sec:supp_mlp_hyper} and~\ref{sec:supp_mricore}.}

\begin{wraptable}[15]{r}{0.53\textwidth}
\centering
\vspace{-0.5cm}
\caption{
\rev{\textbf{MRI$\to$US.}} Ablation on cost terms. The combination of all three terms  results in the most consistent gains. \rev{USFM uses Ridge alignment; SAM-Med2d reports one MLP run per subset.}
}
\label{tab:cost_ablation}
\footnotesize
\setlength{\tabcolsep}{3pt}
\sisetup{
  table-format = 2.1,
  table-number-alignment = center,
  detect-weight = true,
  mode = text
}
\resizebox{1.03\linewidth}{!}{%
\begin{tabular}{
    c c c
    c c
    c c
  }
\toprule
\multicolumn{3}{c}{Active cost terms}
  & \multicolumn{2}{c}{USFM~\cite{jiao2024usfm}}
  & \multicolumn{2}{c}{SAM-Med2d~\cite{cheng2023sam}} \\
\cmidrule(lr){1-3}
\cmidrule(lr){4-5}
\cmidrule(lr){6-7}
$D^\text{time}$ & $\rev{D^\text{proto}}$ & $D^\text{space}$
  & GA\,$\downarrow$
  & Cls.\,$\uparrow$
  & GA\,$\downarrow$
  & Cls.\,$\uparrow$ \\
\midrule

\xmark & \xmark & \xmark
  & \metricbase{21.9}
  & \metricbase{61.9}
  & \metricbase{7.7}
  & \metricbase{83.3} \\

\midrule

\xmark & \xmark & \cmark
  & \metric{35.7}{\badup{13.8}}
  & \metric{\bfseries 76.2}{\goodup{14.3}}
  & \metric{\rev{28.5}}{\badup{\rev{20.8}}}
  & \metric{\rev{90.5}}{\goodup{\rev{7.2}}} \\

\xmark & \cmark & \xmark
  & \metric{20.3}{\gooddown{1.6}}
  & \metric{52.4}{\baddown{9.5}}
  & \metric{\rev{8.9}}{\badup{\rev{1.2}}}
  & \metric{\rev{90.5}}{\goodup{\rev{7.2}}} \\

\xmark & \cmark & \cmark
  & \metric{22.9}{\badup{1.0}}
  & \metric{50.0}{\baddown{11.9}}
  & \metric{\rev{8.6}}{\badup{\rev{0.9}}}
  & \metric{\rev{\bfseries 95.2}}{\goodup{\rev{11.9}}} \\

\cmark & \xmark & \xmark
  & \metric{\bfseries 12.6}{\gooddown{9.3}}
  & \metric{33.3}{\baddown{28.6}}
  & \metric{\rev{\bfseries 6.8}}{\gooddown{\rev{0.9}}}
  & \metric{\rev{64.3}}{\baddown{\rev{19.0}}} \\

\cmark & \xmark & \cmark
  & \metric{\underline{12.9}}{\gooddown{9.0}}
  & \metric{35.7}{\baddown{26.2}}
  & \metric{\rev{8.6}}{\badup{\rev{0.9}}}
  & \rev{\metricbase{83.3}} \\

\cmark & \cmark & \xmark
  & \metric{15.4}{\gooddown{6.5}}
  & \metric{\rev{66.7}}{\goodup{\rev{4.8}}}
  & \metric{\rev{8.2}}{\badup{\rev{0.5}}}
  & \metric{\rev{\underline{92.9}}}{\goodup{\rev{9.6}}} \\

\rowcolor{blue!8}
\cmark & \cmark & \cmark
  & \metric{17.4}{\gooddown{4.5}}
  & \metric{\underline{69.0}}{\goodup{7.1}}
  & \metric{\rev{\underline{7.9}}}{\badup{\rev{0.2}}}
  & \metric{\rev{88.1}}{\goodup{\rev{4.8}}} \\

\bottomrule
\end{tabular}
}
\end{wraptable}

\minisection{Cost Terms.}
We ablate the contribution of each term in the composite cost matrix $\tilde{C}$ by
evaluating all $2^3 = 8$ subsets of $\{D^{\text{time}}, \rev{D^{\text{proto}}}, D^{\text{space}}\}$
under Ridge alignment on USFM and MLP alignment on \rev{SAM-Med2d}. Results are reported in Tab.~\ref{tab:cost_ablation}.
The results reveal a clear complementarity among the three terms. $D^{\text{time}}$ alone
drives strong gestational age regression but collapses plane classification, as it matches
scans across anatomical planes.
\rev{Conversely, $D^{\text{space}}$ alone preserves plane classification but loses the temporal signal, sharply increasing the GA error. Adding plane information to $D^{\text{time}}$ recovers plane accuracy, through $D^{\text{proto}}$ on both backbones and through $D^{\text{space}}$ on SAM-Med2d. The full cost matrix is the only combination that improves plane classification on both backbones while improving GA on USFM and keeping it within $0.2$ days of Zero-Shot on SAM-Med2d,} confirming that each component captures a distinct and complementary aspect of
cross-modal correspondence.

\minisection{Pairing Strategies.}
\rev{To isolate the contribution of the spatiotemporal pairing, we replace the \method transport plan with three metadata-agnostic strategies and evaluate them under the same Ridge, MLP and CCA alignment models used in the main experiments: Random, in which pairing assigns each source sample to a target sample drawn uniformly at random, providing a lower bound on alignment quality; Cosine pairing, which greedily matches each source sample to its nearest neighbor in the target embedding space by cosine similarity, without accounting for the global distribution of assignments; and OT-Cosine, the same entropic OT and barycentric projection as \method but with the cosine distance between raw embeddings as the only cost, which separates the entropic relaxation from the metadata-driven cost, resulting in a fully metadata-free baseline.
Tab.~\ref{tab:mri2us_ablation_pairing} reports results for all four backbones. The three alternative pairing strategies degrade performance relative to Zero-Shot in most settings, and with Ridge and MLP alignment OT-Cosine has a higher GA error than \method on every backbone and a lower plane accuracy in all but two settings (USFM and Echocare with MLP), indicating that most of the gain comes from the spatiotemporal cost rather than from the entropic relaxation alone.}

  \begin{table}[t]
  \centering
  \caption{%
  \rev{\textbf{MRI$\to$US.}} Ablation of pairing strategies without spatiotemporal metadata. \rev{Random pairs each source sample with a target sample drawn uniformly at random, Cosine greedily
  matches each source to its nearest target by cosine similarity in raw embedding space, and OT-Cosine uses the \method entropic OT with the cosine distance between
  raw embeddings as the only cost. All three underperform Zero-Shot in most settings. MLP rows report mean $\pm$ std over 3 seeds; for Random, the seed also changes
  the random pairing. \method rows are reproduced from Tab.~\ref{tab:self_distill_mri2us}.}
  }
  \label{tab:mri2us_ablation_pairing}
  \small
  \resizebox{\textwidth}{!}{%
  \begin{tabular}{l l cc cc cc cc}
  \toprule
  \multicolumn{2}{l}{\multirow{2}{*}{\rev{\textbf{MRI$\to$US}}}} & \multicolumn{4}{c}{Unimodal (US)} & \multicolumn{4}{c}{Multimodal}  \\
  \cmidrule(lr){3-6} \cmidrule(lr){7-10}
  & & \multicolumn{2}{c}{USFM~\cite{jiao2024usfm}} & \multicolumn{2}{c}{Echocare~\cite{zhang2025fully}} & \multicolumn{2}{c}{SAM-Med2d~\cite{cheng2023sam}} &
  \multicolumn{2}{c}{BioMedParse~\cite{zhao2024biomedparse}} \\
  \cmidrule(lr){3-4} \cmidrule(lr){5-6} \cmidrule(lr){7-8} \cmidrule(lr){9-10}
  Pairing & Alignment & GA Reg.\,$\downarrow$ & Plane Cls.\,$\uparrow$ & GA Reg.\,$\downarrow$ & Plane Cls.\,$\uparrow$ & GA Reg.\,$\downarrow$ & Plane
  Cls.\,$\uparrow$ & GA Reg.\,$\downarrow$ & Plane Cls.\,$\uparrow$ \\
  \midrule
  Zero-Shot & --- & 21.9 & 61.9 & 10.3 & 81.0 & \textbf{7.7} & 83.3 & \underline{6.5} & \textbf{85.7} \\
  \midrule
  \multirow{3}{*}{Random}
    & CCA & 28.9 & 35.7 & 36.3 & 38.1 & 28.8 & 40.5 & 31.3 & 47.6 \\
   & Ridge & 32.0 & 42.9 & 22.2 & 66.7 & 29.2 & 54.8 & 20.4 & 61.9 \\
   & MLP & \rev{32.3{\scriptsize$\pm$3.1}} & \rev{45.2{\scriptsize$\pm$10.4}} & \rev{20.2{\scriptsize$\pm$0.7}} & \rev{81.0{\scriptsize$\pm$6.3}} &
  \rev{21.8{\scriptsize$\pm$1.6}} & \rev{71.4{\scriptsize$\pm$2.4}} & \rev{18.6{\scriptsize$\pm$1.8}} & \rev{63.5{\scriptsize$\pm$5.0}}\\
  \midrule
  \multirow{3}{*}{Cosine}
    & CCA & 25.0 & 35.7 & 23.3 & 54.8 & 29.8 & 40.5 & 20.3 & 50.0 \\
   & Ridge & 25.0 & 40.5 & 31.2 & 66.7 & 27.9 & 64.3 & 14.9 & 50.0 \\
   & MLP & \rev{26.0{\scriptsize$\pm$0.2}} & \rev{36.5{\scriptsize$\pm$1.4}} & \rev{20.8{\scriptsize$\pm$1.4}} & \rev{\underline{84.9{\scriptsize$\pm$2.7}}} &
  \rev{16.2{\scriptsize$\pm$0.1}} & \rev{84.9{\scriptsize$\pm$1.4}} & \rev{11.8{\scriptsize$\pm$0.8}} & \rev{57.9{\scriptsize$\pm$5.0}}\\
  \midrule
  \multirow{3}{*}{\rev{OT-Cosine}}
    & CCA & \rev{25.3} & \rev{59.5} & \rev{22.5} & \rev{57.1} & \rev{44.1} & \rev{38.1} & \rev{22.5} & \rev{40.5}\\
   & Ridge & \rev{24.6} & \rev{64.3} & \rev{16.6} & \rev{81.0} & \rev{22.8} & \rev{71.4} & \rev{12.8} & \rev{59.5}\\
   & MLP & \rev{22.8{\scriptsize$\pm$0.1}} & \rev{\textbf{69.0{\scriptsize$\pm$0.0}}} & \rev{15.7{\scriptsize$\pm$2.2}} & \rev{\underline{84.9{\scriptsize$\pm$2.7}}} &
  \rev{10.5{\scriptsize$\pm$0.6}} & \rev{84.1{\scriptsize$\pm$5.0}} & \rev{12.5{\scriptsize$\pm$1.4}} & \rev{69.8{\scriptsize$\pm$1.4}}\\
  \midrule
  \rowcolor{blue!8}  & CCA & \rev{\underline{17.7}} & \rev{64.3} & \rev{11.9} & \rev{81.0} & \rev{28.3} & \rev{50.0} & \rev{15.2} & \rev{54.8}\\
  \rowcolor{blue!8} \textbf{\method (Ours)} & Ridge & \textbf{17.4} & \textbf{69.0} & \textbf{9.1} & \textbf{90.5} & 9.0 & \underline{88.1} & 6.9 & \textbf{85.7} \\
  \rowcolor{blue!8}  & MLP & \rev{19.9{\scriptsize$\pm$0.1}} & \rev{\underline{65.1{\scriptsize$\pm$1.4}}} & \rev{\underline{9.7{\scriptsize$\pm$0.5}}} &
  \rev{84.1{\scriptsize$\pm$1.4}} & \rev{\underline{7.9{\scriptsize$\pm$0.3}}} & \rev{\textbf{89.7{\scriptsize$\pm$2.7}}} & \rev{\textbf{6.0{\scriptsize$\pm$0.6}}} &
  \rev{\underline{84.1{\scriptsize$\pm$1.4}}}\\
  \bottomrule
  \end{tabular}}
  \end{table}

\raggedbottom
\section{Conclusions}
\label{sec:conclusions}
We introduced \method, a framework for establishing cross-modal correspondence between fetal MRI and fetal US without requiring subject-level paired acquisitions. By leveraging two routinely available pieces of clinical metadata, gestational age and standard diagnostic plane, as the cost function of an entropic Optimal Transport problem, \method derives soft correspondences between independent unimodal cohorts. These correspondences are sufficient to learn a lightweight \rev{alignment module} on top of frozen visual backbones.
This alignment enables effective knowledge transfer across modalities. Applied to US-only foundation models, \method allows representations learned exclusively from ultrasound to generalize to MRI, a previously unseen modality, substantially improving downstream performance. Additionally, \method further refines representations of multimodal backbones already exposed to both MRI and US, yielding \rev{gains on at least one task for every multimodal backbone} even in near-saturated settings. \rev{In the reverse US$\to$MRI direction, \method preserves the strong Zero-Shot representations of all five backbones.} We additionally show that multimodal improvements can \rev{often} be distilled from a stronger teacher into a lighter student \rev{via a single intra-modal regression.}

\rev{Our results suggest that routinely collected clinical metadata can serve as a cross-modal supervision signal when subject-level pairs are unavailable, a principle that may apply beyond fetal imaging.}  
  
\minisection{Limitations.} 
Our framework lacks a direct comparison with paired-data methods, as no paired fetal US--MRI dataset is available; matched atlas pairs could serve as a proxy in future work. Using single MRI slices may also be suboptimal, and incorporating neighboring slices could better exploit the 3D context. US gestational ages are Hadlock-estimated from biometric landmarks rather than independently recorded, introducing uncertainty into the GA regression target and potentially sharing anatomical information with the input image.
Finally, \method is evaluated only on neurotypical fetuses, and its behavior in the presence of anatomical anomalies remains an open question.

\begin{ack}

Niccol\`o Biondi acknowledges the support of the EU Horizon Europe projects
ELIAS (No.~101120237) and ELLIOT (No.~101214398), and of the FIS project
GUIDANCE (No.~FIS2023-03251).

This research was supported by the National Plan for Complementary Investments
to the PNRR Hub Life Science-Digital Health
(PNC-E3-2022-23683267, DHEAL COM).
This publication reflects only the authors' view and the Italian Ministry of
Health is not responsible for any use that may be made of the information it
contains.

\end{ack}

{
\small
\bibliographystyle{unsrtnat}
\bibliography{bib}

\begin{thebibliography}{54}
\providecommand{\natexlab}[1]{#1}
\providecommand{\url}[1]{\texttt{#1}}
\expandafter\ifx\csname urlstyle\endcsname\relax
  \providecommand{\doi}[1]{doi: #1}\else
  \providecommand{\doi}{doi: \begingroup \urlstyle{rm}\Url}\fi

\bibitem[Azad et~al.(2023)Azad, Azad, Eskandari, Bozorgpour, Kazerouni, Rekik,
  and Merhof]{azad2023foundational}
Bobby Azad, Reza Azad, Sania Eskandari, Afshin Bozorgpour, Amirhossein
  Kazerouni, Islem Rekik, and Dorit Merhof.
\newblock Foundational models in medical imaging: A comprehensive survey and
  future vision.
\newblock \emph{arXiv preprint arXiv:2310.18689}, 2023.

\bibitem[Gholipour et~al.(2017)Gholipour, Rollins, Velasco-Annis, Ouaalam,
  Akhondi-Asl, Afacan, Ortinau, Clancy, Limperopoulos, Yang,
  et~al.]{gholipour2017normative}
Ali Gholipour, Caitlin~K Rollins, Clemente Velasco-Annis, Abdelhakim Ouaalam,
  Alireza Akhondi-Asl, Onur Afacan, Cynthia~M Ortinau, Sean Clancy, Catherine
  Limperopoulos, Edward Yang, et~al.
\newblock A normative spatiotemporal {MRI} atlas of the fetal brain for
  automatic segmentation and analysis of early brain growth.
\newblock \emph{Scientific reports}, 7\penalty0 (1):\penalty0 476, 2017.

\bibitem[Malinger et~al.(2020)Malinger, Paladini, Haratz, Monteagudo, Pilu, and
  Timor-Tritsch]{malinger2020isuog}
G~Malinger, D~Paladini, K~K Haratz, A~Monteagudo, GL~Pilu, and
  IE~Timor-Tritsch.
\newblock {ISUOG} practice guidelines (updated): sonographic examination of the
  fetal central nervous system. {P}art 1: performance of screening examination
  and indications for targeted neurosonography.
\newblock \emph{Ultrasound in obstetrics \& gynecology: the official journal of
  the International Society of Ultrasound in Obstetrics and Gynecology},
  56\penalty0 (3):\penalty0 476--484, 2020.

\bibitem[Jiao et~al.(2024)Jiao, Zhou, Li, Xia, Huang, Huang, Wang, Zhang, Zhou,
  Wang, et~al.]{jiao2024usfm}
Jing Jiao, Jin Zhou, Xiaokang Li, Menghua Xia, Yi~Huang, Lihong Huang, Na~Wang,
  Xiaofan Zhang, Shichong Zhou, Yuanyuan Wang, et~al.
\newblock {USFM}: A universal ultrasound foundation model generalized to tasks
  and organs towards label efficient image analysis.
\newblock \emph{Medical image analysis}, 96:\penalty0 103202, 2024.

\bibitem[Zhang et~al.(2025)Zhang, Wu, Zhao, Chen, Li, Zhu, Zhao, Yuan, Yang,
  Qiu, et~al.]{zhang2025fully}
Hongyuan Zhang, Yuheng Wu, Mingyang Zhao, Zhiwei Chen, Rebecca Li, Fei Zhu,
  Haohan Zhao, Xiaohua Yuan, Meng Yang, Chunli Qiu, et~al.
\newblock A fully open and generalizable foundation model for ultrasound
  clinical applications.
\newblock \emph{arXiv preprint arXiv:2509.11752}, 2025.

\bibitem[Radhachandran et~al.(2026)Radhachandran, Ivezi{\'c}, Athreya,
  Anilkumar, Arnold, and Speier]{radhachandran2026us}
Ashwath Radhachandran, Vedrana Ivezi{\'c}, Shreeram Athreya, Ronit Anilkumar,
  Corey~W Arnold, and William Speier.
\newblock {US-JEPA}: A joint embedding predictive architecture for medical
  ultrasound.
\newblock \emph{arXiv preprint arXiv:2602.19322}, 2026.

\bibitem[Ambsdorf et~al.(2025)Ambsdorf, Munk, Llambias, Christensen, Mikolaj,
  Balestriero, Tolsgaard, Feragen, and Nielsen]{ambsdorf2025general}
Jakob Ambsdorf, Asbj{\o}rn Munk, Sebastian Llambias, Anders~N Christensen,
  Kamil Mikolaj, Randall Balestriero, Martin~G Tolsgaard, Aasa Feragen, and
  Mads Nielsen.
\newblock General methods make great domain-specific foundation models: A
  case-study on fetal ultrasound.
\newblock In \emph{International Conference on Medical Image Computing and
  Computer-Assisted Intervention}, pages 271--281. Springer, 2025.

\bibitem[Dong et~al.(2025)Dong, Chen, Gu, Konz, Chen, Li, and
  Mazurowski]{dong2025mri}
Haoyu Dong, Yuwen Chen, Hanxue Gu, Nicholas Konz, Yaqian Chen, Qihang Li, and
  Maciej~A Mazurowski.
\newblock {MRI-CORE}: a foundation model for magnetic resonance imaging.
\newblock \emph{arXiv preprint arXiv:2506.12186}, 2025.

\bibitem[Cheng et~al.(2023)Cheng, Ye, Deng, Chen, Li, Wang, Su, Huang, Chen,
  Jiang, et~al.]{cheng2023sam}
Junlong Cheng, Jin Ye, Zhongying Deng, Jianpin Chen, Tianbin Li, Haoyu Wang,
  Yanzhou Su, Ziyan Huang, Jilong Chen, Lei Jiang, et~al.
\newblock {SAM-Med2D}.
\newblock \emph{arXiv preprint arXiv:2308.16184}, 2023.

\bibitem[Zhao et~al.(2025)Zhao, Gu, Yang, Usuyama, Lee, Kiblawi, Naumann, Gao,
  Crabtree, Abel, et~al.]{zhao2024biomedparse}
Theodore Zhao, Yu~Gu, Jianwei Yang, Naoto Usuyama, Ho~Hin Lee, Sid Kiblawi,
  Tristan Naumann, Jianfeng Gao, Angela Crabtree, Jacob Abel, et~al.
\newblock A foundation model for joint segmentation, detection and recognition
  of biomedical objects across nine modalities.
\newblock \emph{Nature Methods}, 22:\penalty0 166--176, 2025.
\newblock \doi{10.1038/s41592-024-02499-w}.

\bibitem[Namburete et~al.(2023)Namburete, Papie{\.z}, Fernandes, Wyburd, Hesse,
  Moser, Ismail, Gunier, Squier, Ohuma, et~al.]{namburete2023normative}
Ana~IL Namburete, Bart{\l}omiej~W Papie{\.z}, Michelle Fernandes, Madeleine~K
  Wyburd, Linde~S Hesse, Felipe~A Moser, Leila~Cheikh Ismail, Robert~B Gunier,
  Waney Squier, Eric~O Ohuma, et~al.
\newblock Normative spatiotemporal fetal brain maturation with satisfactory
  development at 2 years.
\newblock \emph{Nature}, 623\penalty0 (7985):\penalty0 106--114, 2023.

\bibitem[Cuturi(2013)]{cuturi2013sinkhorn}
Marco Cuturi.
\newblock {S}inkhorn distances: Lightspeed computation of optimal transport.
\newblock \emph{Advances in neural information processing systems}, 26, 2013.

\bibitem[Zhu et~al.(2017)Zhu, Park, Isola, and Efros]{zhu2017unpaired}
Jun-Yan Zhu, Taesung Park, Phillip Isola, and Alexei~A Efros.
\newblock Unpaired image-to-image translation using cycle-consistent
  adversarial networks.
\newblock In \emph{Proceedings of the IEEE international conference on computer
  vision}, pages 2223--2232, 2017.

\bibitem[Chen et~al.(2023)Chen, Chen, Wee, Dekker, and Bermejo]{chen2023deep}
Junhua Chen, Shenlun Chen, Leonard Wee, Andre Dekker, and Inigo Bermejo.
\newblock Deep learning based unpaired image-to-image translation applications
  for medical physics: a systematic review.
\newblock \emph{Physics in Medicine \& Biology}, 68\penalty0 (5):\penalty0
  05TR01, 2023.

\bibitem[Silverstein et~al.(2025)Silverstein, Leibowitz, Beloosesky, and
  Azhari]{silverstein2025translation}
Naomi Silverstein, Efrat Leibowitz, Ron Beloosesky, and Haim Azhari.
\newblock Translation of fetal brain ultrasound images into pseudo-{MRI} images
  using artificial intelligence.
\newblock \emph{arXiv preprint arXiv:2504.02408}, 2025.

\bibitem[Jiao et~al.(2020)Jiao, Namburete, Papageorghiou, and
  Noble]{jiao2020self}
Jianbo Jiao, Ana~IL Namburete, Aris~T Papageorghiou, and J~Alison Noble.
\newblock Self-supervised ultrasound to {MRI} fetal brain image synthesis.
\newblock \emph{IEEE Transactions on Medical Imaging}, 39\penalty0
  (12):\penalty0 4413--4424, 2020.

\bibitem[Sun et~al.(2017)Sun, Feng, and Saenko]{sun2017correlation}
Baochen Sun, Jiashi Feng, and Kate Saenko.
\newblock Correlation alignment for unsupervised domain adaptation.
\newblock In \emph{Domain adaptation in computer vision applications}, pages
  153--171. Springer, 2017.

\bibitem[Dou et~al.(2018)Dou, Ouyang, Chen, Chen, and
  Heng]{dou2018unsupervised}
Qi~Dou, Cheng Ouyang, Cheng Chen, Hao Chen, and Pheng-Ann Heng.
\newblock Unsupervised cross-modality domain adaptation of {ConvNets} for
  biomedical image segmentations with adversarial loss.
\newblock In \emph{Proceedings of the Twenty-Seventh International Joint
  Conference on Artificial Intelligence ({IJCAI})}, pages 691--697, 2018.
\newblock \doi{10.24963/ijcai.2018/96}.

\bibitem[Courty et~al.(2017)Courty, Flamary, Tuia, and
  Rakotomamonjy]{courty2016optimal}
Nicolas Courty, R{\'e}mi Flamary, Devis Tuia, and Alain Rakotomamonjy.
\newblock Optimal transport for domain adaptation.
\newblock \emph{IEEE transactions on pattern analysis and machine
  intelligence}, 39\penalty0 (9):\penalty0 1853--1865, 2017.

\bibitem[Alvarez-Melis and Jaakkola(2018)]{alvarez2018gromov}
David Alvarez-Melis and Tommi Jaakkola.
\newblock {Gromov-Wasserstein} alignment of word embedding spaces.
\newblock In \emph{Proceedings of the 2018 conference on empirical methods in
  natural language processing}, pages 1881--1890, 2018.

\bibitem[Demetci et~al.(2022)Demetci, Santorella, Sandstede, Noble, and
  Singh]{demetci2022scot}
Pinar Demetci, Rebecca Santorella, Bj{\"o}rn Sandstede, William~Stafford Noble,
  and Ritambhara Singh.
\newblock {SCOT}: single-cell multi-omics alignment with optimal transport.
\newblock \emph{Journal of computational biology}, 29\penalty0 (1):\penalty0
  3--18, 2022.

\bibitem[Peyr{\'e} and Cuturi(2019)]{peyre2019computational}
Gabriel Peyr{\'e} and Marco Cuturi.
\newblock \emph{Computational optimal transport: With applications to data
  science}.
\newblock Now Foundations and Trends, 2019.

\bibitem[Fiorentino et~al.(2025)Fiorentino, Moccia, Cosmo, Frontoni, Giovanola,
  and Tiribelli]{fiorentino2025uncovering}
Maria~Chiara Fiorentino, Sara Moccia, Mariachiara~Di Cosmo, Emanuele Frontoni,
  Benedetta Giovanola, and Simona Tiribelli.
\newblock Uncovering ethical biases in publicly available fetal ultrasound
  datasets.
\newblock \emph{npj Digital Medicine}, 8\penalty0 (1):\penalty0 355, 2025.

\bibitem[Vahedifard et~al.(2023)Vahedifard, Adepoju, Supanich, Ai, Liu, Kocak,
  Marathu, and Byrd]{vahedifard2023review}
Farzan Vahedifard, Jubril~O Adepoju, Mark Supanich, Hua~Asher Ai, Xuchu Liu,
  Mehmet Kocak, Kranthi~K Marathu, and Sharon~E Byrd.
\newblock Review of deep learning and artificial intelligence models in fetal
  brain magnetic resonance imaging.
\newblock \emph{World journal of clinical cases}, 11\penalty0 (16):\penalty0
  3725, 2023.

\bibitem[Maani et~al.(2026)Maani, Saeed, Saleem, Farooq, Alasmawi, Diehl,
  Mohammad, Waring, Valappil, Bricker, and Yaqub]{maani2025fetalclip}
Fadillah Maani, Numan Saeed, Tausifa~Jan Saleem, Zaid Farooq, Hussain Alasmawi,
  Werner Diehl, Ameera Mohammad, Gareth Waring, Saudabi Valappil, Leanne
  Bricker, and Mohammad Yaqub.
\newblock {FetalCLIP}: A visual-language foundation model for fetal ultrasound
  image analysis.
\newblock \emph{npj Digital Medicine}, 2026.
\newblock \doi{10.1038/s41746-026-02907-9}.

\bibitem[Fuerst et~al.(2014)Fuerst, Wein, M{\"u}ller, and
  Navab]{fuerst2014automatic}
Bernhard Fuerst, Wolfgang Wein, Markus M{\"u}ller, and Nassir Navab.
\newblock Automatic ultrasound--{MRI} registration for neurosurgery using the
  {2D} and {3D} {LC2} metric.
\newblock \emph{Medical image analysis}, 18\penalty0 (8):\penalty0 1312--1319,
  2014.

\bibitem[Darzi and Bocklitz(2024)]{darzi2024review}
Fatemehzahra Darzi and Thomas Bocklitz.
\newblock A review of medical image registration for different modalities.
\newblock \emph{Bioengineering}, 11\penalty0 (8):\penalty0 786, 2024.

\bibitem[Marami et~al.(2017)Marami, Salehi, Afacan, Scherrer, Rollins, Yang,
  Estroff, Warfield, and Gholipour]{marami2017temporal}
Bahram Marami, Seyed Sadegh~Mohseni Salehi, Onur Afacan, Benoit Scherrer,
  Caitlin~K Rollins, Edward Yang, Judy~A Estroff, Simon~K Warfield, and Ali
  Gholipour.
\newblock Temporal slice registration and robust diffusion-tensor
  reconstruction for improved fetal brain structural connectivity analysis.
\newblock \emph{NeuroImage}, 156:\penalty0 475--488, 2017.

\bibitem[Wyburd et~al.(2025)Wyburd, Dinsdale, Kyriakopoulou, Venturini, Wright,
  Uus, Matthew, Skelton, Z{\"o}llei, Hajnal, et~al.]{wyburd2025cross}
Madeleine~K Wyburd, Nicola~K Dinsdale, Vanessa Kyriakopoulou, Lorenzo
  Venturini, Robert Wright, Alena Uus, Jacqueline Matthew, Emily Skelton, Lilla
  Z{\"o}llei, Joseph Hajnal, et~al.
\newblock Cross-modality comparison of fetal brain phenotypes: Insights from
  short-interval second-trimester {MRI} and ultrasound imaging.
\newblock \emph{Human Brain Mapping}, 46\penalty0 (14):\penalty0 e70349, 2025.

\bibitem[Calixto et~al.(2024)Calixto, Machado-Rivas, Karimi, Velasco-Annis,
  Cortes-Albornoz, Afacan, Warfield, Gholipour, and
  Jaimes]{calixto2024population}
Camilo Calixto, Fedel Machado-Rivas, Davood Karimi, Clemente Velasco-Annis,
  Maria~Camila Cortes-Albornoz, Onur Afacan, Simon~K Warfield, Ali Gholipour,
  and Camilo Jaimes.
\newblock Population atlas analysis of emerging brain structural connections in
  the human fetus.
\newblock \emph{Journal of Magnetic Resonance Imaging}, 60\penalty0
  (1):\penalty0 152--160, 2024.

\bibitem[Bastiaansen et~al.(2026)Bastiaansen, Rousian, Koning, Niessen,
  de~Bakker, Steegers-Theunissen, and Klein]{bastiaansen20264d}
Wietske~AP Bastiaansen, Melek Rousian, Anton~HJ Koning, Wiro~J Niessen,
  Bernadette~S de~Bakker, R{\'e}gine~PM Steegers-Theunissen, and Stefan Klein.
\newblock The {4D} human embryonic brain atlas: Spatiotemporal atlas generation
  for rapid anatomical changes.
\newblock \emph{Computerized Medical Imaging and Graphics}, page 102702, 2026.

\bibitem[Di~Vece et~al.(2026)Di~Vece, Mao, Avisdris, Dromey, Napolitano,
  Ben~Bashat, Vasconcelos, Stoyanov, Joskowicz, and Bano]{di2025multi}
Chiara Di~Vece, Zhehua Mao, Netanell Avisdris, Brian Dromey, Raffaele
  Napolitano, Dafna Ben~Bashat, Francisco Vasconcelos, Danail Stoyanov, Leo
  Joskowicz, and Sophia Bano.
\newblock A multicentre benchmark dataset for comprehensive landmark-based
  fetal ultrasound biometry.
\newblock \emph{Scientific Reports}, 16:\penalty0 17405, 2026.
\newblock \doi{10.1038/s41598-026-47854-3}.

\bibitem[Burgos-Artizzu et~al.(2020)Burgos-Artizzu, Coronado-Guti{\'e}rrez,
  Valenzuela-Alcaraz, Bonet-Carne, Eixarch, Crispi, and
  Gratac{\'o}s]{burgos2020evaluation}
Xavier~P Burgos-Artizzu, David Coronado-Guti{\'e}rrez, Brenda
  Valenzuela-Alcaraz, Elisenda Bonet-Carne, Elisenda Eixarch, Fatima Crispi,
  and Eduard Gratac{\'o}s.
\newblock Evaluation of deep convolutional neural networks for automatic
  classification of common maternal fetal ultrasound planes.
\newblock \emph{Scientific Reports}, 10\penalty0 (1):\penalty0 10200, 2020.

\bibitem[van~den Heuvel et~al.(2018)van~den Heuvel, de~Bruijn, de~Korte, and
  Ginneken]{van2018automated}
Thomas~LA van~den Heuvel, Dagmar de~Bruijn, Chris~L de~Korte, and Bram~van
  Ginneken.
\newblock Automated measurement of fetal head circumference using {2D}
  ultrasound images.
\newblock \emph{PloS one}, 13\penalty0 (8):\penalty0 e0200412, 2018.

\bibitem[Bano et~al.(2021)Bano, Dromey, Vasconcelos, Napolitano, David,
  Peebles, and Stoyanov]{bano2021autofb}
Sophia Bano, Brian Dromey, Francisco Vasconcelos, Raffaele Napolitano, Anna~L
  David, Donald~M Peebles, and Danail Stoyanov.
\newblock {AutoFB}: automating fetal biometry estimation from standard
  ultrasound planes.
\newblock In \emph{International Conference on Medical Image Computing and
  Computer-Assisted Intervention}, pages 228--238. Springer, 2021.

\bibitem[Hadlock et~al.(1984)Hadlock, Deter, Harrist, and
  Park]{hadlock1984estimating}
Frank~P Hadlock, Russell~L Deter, Ronald~B Harrist, and SK~Park.
\newblock Estimating fetal age: computer-assisted analysis of multiple fetal
  growth parameters.
\newblock \emph{Radiology}, 152\penalty0 (2):\penalty0 497--501, 1984.

\bibitem[Zalevskyi et~al.(2026)Zalevskyi, Sanchez, Kaandorp, Roulet,
  Fajardo-Rojas, Li, Hutter, Li, Barkovich, Ji, et~al.]{zalevskyi2026advances}
Vladyslav Zalevskyi, Thomas Sanchez, Misha Kaandorp, Margaux Roulet, Diego
  Fajardo-Rojas, Liu Li, Jana Hutter, Hongwei~Bran Li, Matthew~J Barkovich, Hui
  Ji, et~al.
\newblock Advances in automated fetal brain mri segmentation and biometry:
  insights from the {FeTA} 2024 challenge.
\newblock \emph{Medical image analysis}, page 103941, 2026.

\bibitem[Gong et~al.(2025)Gong, Liu, Wang, Liu, Wan, Shi, and
  Li]{gong2025fetal}
Haifan Gong, Huixian Liu, Yitao Wang, Xiaoling Liu, Xiang Wan, Qiao Shi, and
  Haofeng Li.
\newblock Fetal cerebellum landmark detection based on {3D} {MRI}: Method and
  benchmark.
\newblock \emph{IEEE Journal of Biomedical and Health Informatics}, 2025.

\bibitem[Hardoon et~al.(2004)Hardoon, Szedmak, and
  Shawe-Taylor]{hardoon2004canonical}
David~R Hardoon, Sandor Szedmak, and John Shawe-Taylor.
\newblock Canonical correlation analysis: An overview with application to
  learning methods.
\newblock \emph{Neural computation}, 16\penalty0 (12):\penalty0 2639--2664,
  2004.

\bibitem[Dorent et~al.(2026)Dorent, Haouchine, Golby, Frisken, Kapur, and
  Wells]{dorent2026unified}
Reuben Dorent, Nazim Haouchine, Alexandra Golby, Sarah Frisken, Tina Kapur, and
  William Wells.
\newblock Unified cross-modal medical image synthesis with hierarchical mixture
  of product-of-experts.
\newblock \emph{IEEE Transactions on Pattern Analysis and Machine
  Intelligence}, 48\penalty0 (2):\penalty0 1641--1656, 2026.
\newblock \doi{10.1109/TPAMI.2025.3616632}.

\bibitem[Tzeng et~al.(2017)Tzeng, Hoffman, Saenko, and
  Darrell]{tzeng2017adversarial}
Eric Tzeng, Judy Hoffman, Kate Saenko, and Trevor Darrell.
\newblock Adversarial discriminative domain adaptation.
\newblock In \emph{Proceedings of the IEEE Conference on Computer Vision and
  Pattern Recognition (CVPR)}, pages 2962--2971, 2017.

\bibitem[Govaert et~al.(2020)Govaert, Triulzi, and
  Dudink]{govaert2020developing}
Paul Govaert, Fabio Triulzi, and Jeroen Dudink.
\newblock The developing brain by trimester.
\newblock \emph{Handbook of Clinical Neurology}, 171:\penalty0 245--289, 2020.

\bibitem[Moltoni et~al.(2021)Moltoni, Talenti, and Righini]{moltoni2021brain}
Giulia Moltoni, Giacomo Talenti, and Andrea Righini.
\newblock Brain fetal neuroradiology: a beginner’s guide.
\newblock \emph{Translational pediatrics}, 10\penalty0 (4):\penalty0 1065,
  2021.

\bibitem[Rakic(1972)]{rakic1972mode}
Pasko Rakic.
\newblock Mode of cell migration to the superficial layers of fetal monkey
  neocortex.
\newblock \emph{Journal of Comparative Neurology}, 145\penalty0 (1):\penalty0
  61--83, 1972.

\bibitem[Subramanian et~al.(2020)Subramanian, Calcagnotto, and
  Paredes]{subramanian2020cortical}
Lakshmi Subramanian, Maria~Elisa Calcagnotto, and Mercedes~F Paredes.
\newblock Cortical malformations: lessons in human brain development.
\newblock \emph{Frontiers in cellular neuroscience}, 13:\penalty0 576, 2020.

\bibitem[Chi et~al.(1977)Chi, Dooling, and Gilles]{chi1977gyral}
Je~G Chi, Elizabeth~C Dooling, and Floyd~H Gilles.
\newblock Gyral development of the human brain.
\newblock \emph{Annals of Neurology: Official Journal of the American
  Neurological Association and the Child Neurology Society}, 1\penalty0
  (1):\penalty0 86--93, 1977.

\bibitem[Garel et~al.(2001)Garel, Chantrel, Brisse, Elmaleh, Luton, Oury,
  Sebag, and Hassan]{garel2001fetal}
Catherine Garel, Emmanuel Chantrel, Herv{\'e} Brisse, Monique Elmaleh,
  Dominique Luton, Jean-Fran{\c{c}}ois Oury, Guy Sebag, and Max Hassan.
\newblock Fetal cerebral cortex: normal gestational landmarks identified using
  prenatal {MR} imaging.
\newblock \emph{American Journal of Neuroradiology}, 22\penalty0 (1):\penalty0
  184--189, 2001.

\bibitem[Dubois et~al.(2019)Dubois, Lef{\`e}vre, Angleys, Leroy, Fischer,
  Lebenberg, Dehaene-Lambertz, Borradori-Tolsa, Lazeyras, Hertz-Pannier,
  et~al.]{dubois2019dynamics}
Jessica Dubois, Julien Lef{\`e}vre, Hugo Angleys, Fran{\c{c}}ois Leroy, Clara
  Fischer, Jessica Lebenberg, Ghislaine Dehaene-Lambertz, Cristina
  Borradori-Tolsa, Fran{\c{c}}ois Lazeyras, Lucie Hertz-Pannier, et~al.
\newblock The dynamics of cortical folding waves and prematurity-related
  deviations revealed by spatial and spectral analysis of gyrification.
\newblock \emph{Neuroimage}, 185:\penalty0 934--946, 2019.

\bibitem[Dubois et~al.(2008)Dubois, Benders, Borradori-Tolsa, Cachia, Lazeyras,
  Ha-Vinh~Leuchter, Sizonenko, Warfield, Mangin, and
  H{\"u}ppi]{dubois2008primary}
Jessica Dubois, M~Benders, C~Borradori-Tolsa, Arnaud Cachia, F~Lazeyras,
  R~Ha-Vinh~Leuchter, SV~Sizonenko, SK~Warfield, JF~Mangin, and PS~H{\"u}ppi.
\newblock Primary cortical folding in the human newborn: an early marker of
  later functional development.
\newblock \emph{Brain}, 131\penalty0 (8):\penalty0 2028--2041, 2008.

\bibitem[Hu et~al.(2025)Hu, Zhan, Lu, and Li]{hu2025choroid}
Wei-Xi Hu, Xin Zhan, Dan Lu, and Zhi-Qiang Li.
\newblock Is choroid plexus growth altered in isolated ventriculomegaly on
  fetal neuro-ultrasound?
\newblock \emph{European Radiology}, 35\penalty0 (1):\penalty0 463--473, 2025.

\bibitem[Winter et~al.(2010)Winter, Kennedy, Byrne, and
  Woodward]{winter2010cavum}
Thomas~C Winter, Anne~M Kennedy, Jan Byrne, and Paula~J Woodward.
\newblock The cavum septi pellucidi: why is it important?
\newblock \emph{Journal of Ultrasound in Medicine}, 29\penalty0 (3):\penalty0
  427--444, 2010.

\bibitem[Nagaraj et~al.(2018)Nagaraj, Calvo-Garcia, and
  Kline-Fath]{nagaraj2018abnormalities}
Usha~D Nagaraj, Maria~A Calvo-Garcia, and Beth~M Kline-Fath.
\newblock Abnormalities associated with the cavum septi pellucidi on fetal
  {MRI}: what radiologists need to know.
\newblock \emph{American Journal of Roentgenology}, 210\penalty0 (5):\penalty0
  989--997, 2018.

\bibitem[Scott et~al.(2012)Scott, Hamzelou, Rajagopalan, Habas, Kim, Barkovich,
  Glenn, and Studholme]{scott20123d}
Julia~A Scott, Kia~S Hamzelou, Vidya Rajagopalan, Piotr~A Habas, Kio Kim,
  A~James Barkovich, Orit~A Glenn, and Colin Studholme.
\newblock {3D} morphometric analysis of human fetal cerebellar development.
\newblock \emph{The Cerebellum}, 11\penalty0 (3):\penalty0 761--770, 2012.

\bibitem[Liu et~al.(2011)Liu, Zhang, Lin, Teng, Meng, Yu, Fang, Zang, Li, and
  Liu]{liu2011development}
Fei Liu, Zhonghe Zhang, Xiangtao Lin, Gaojun Teng, Haiwei Meng, Taifei Yu, Fang
  Fang, Fengchao Zang, Zhenping Li, and Shuwei Liu.
\newblock Development of the human fetal cerebellum in the second trimester: a
  post mortem magnetic resonance imaging evaluation.
\newblock \emph{Journal of anatomy}, 219\penalty0 (5):\penalty0 582--588, 2011.

\end{thebibliography}
}

\appendix
\clearpage
\begin{appendices}

\section{Additional Dataset Details}
\label{sec:supp_data}

$\mathcal{D}^{\text{US}}$ contains all the brain images from the \rev{``Multicentre benchmark dataset for comprehensive landmark-based fetal ultrasound biometry''}~\cite{di2025multi}, a curated collection derived from the Fetal Plane dataset~\cite{burgos2020evaluation}, the head circumference dataset~\cite{van2018automated}, and an expanded clinical cohort from University College London Hospital~\cite{bano2021autofb}. It is augmented with non-transthalamic brain images from~\cite{burgos2020evaluation} to cover all three standard diagnostic planes. Since gestational age targets are unavailable in the US datasets, $g^{\text{US}}_i$ is estimated from whole-body biometric measurements using Hadlock's regression formulas~\cite{hadlock1984estimating}. 

Primary fetal biometric parameters - Biparietal Diameter (BPD), Occipitofrontal Diameter (OFD), Transverse Abdominal Diameter (TAD), Anteroposterior Abdominal Diameter (APAD), Femur Length (FL), Head Circumference (HC), and Abdominal Circumference (AC) — are extracted from the 2D spatial landmark annotations provided in~\cite{di2025multi}.To maximize the reliability of gestational-age estimation for each image, we evaluate all valid multi-parameter Hadlock equations supported by the available biometrics and select the prediction with the minimum theoretical standard deviation.

$\mathcal{D}^{\text{MRI}}$ is built from three datasets: FeTa~\cite{zalevskyi2026advances}, LFC~\cite{gong2025fetal}, and the Computational Radiology Laboratory (CRL) normative spatiotemporal MRI atlas~\cite{gholipour2017normative}. Natively, fetal MRI is acquired as overlapping stacks of thick 2D T2-weighted slices, which are integrated via Super-Resolution Reconstruction (SRR) into a high-resolution 3D volume. We leverage already-reconstructed SRR volumes from these three sources.

\subsection{Dataset Statistics}

$\mathcal{D}^{\text{US}}$ comprises $3{,}369$ images from $1{,}744$ unique subjects, covering a gestational age range of $86$--$284$ days ($\approx$12.3--40.6 gestational weeks), with a mean of $163.3 \pm 45.7$ days and a median of $143$ days. 

The distribution of anatomical planes is strongly imbalanced: the transthalamic (TT) plane accounts for the large majority of the data ($2{,}756$ images, $\approx$81.8\%), while the transcerebellar (TC) and transventricular (TV) planes represent $353$ ($\approx$10.5\%) and $260$ ($\approx$7.7\%) images, respectively.
This imbalance reflects the clinical prevalence of TT acquisitions in routine fetal neurosonography practice.
$\mathcal{D}^{\text{US}}$ is partitioned at the \rev{subject} level into three disjoint subsets. 

The assignment to $\mathcal{D}^{\text{US}}_{\text{pair}}$ is driven by temporal compatibility with the MRI cohort: \rev{only US images from $140$ days ($20$ weeks) onward, i.e., close to or within the coverage window of $\mathcal{D}^{\text{MRI}}$ ($147$--$266$ days), are eligible to form cross-modal correspondences}, since US images at younger gestational ages have no MRI counterpart against which a meaningful pairing can be established. Consequently, $\mathcal{D}^{\text{US}}_{\text{pair}}$ is enriched in older subjects (mean GA $176.7$ days, range $140$--$284$ days), whose ages overlap substantially with the MRI coverage window ($147$--$266$ days); the younger end of the US spectrum, with no MRI counterpart against which a meaningful transport cost can be computed, is instead directed to $\mathcal{D}^{\text{US}}_{\text{train}}$ and $\mathcal{D}^{\text{US}}_{\text{test}}$, where it contributes exclusively to downstream probe fitting and evaluation. Note that the two pair sets need not have identical GA distributions: the entropic optimal transport plan \rev{$P$} is computed over their joint support and naturally up-weights source--target pairs that are close in gestational age, handling any residual distributional asymmetry between the modalities. This selection yields a natural 2:4:1 image ratio (pair:train:test), with the remaining non-pair images split 80/20 between train and test. Each source dataset contributes proportionally to all three splits. 

$\mathcal{D}^{\text{US}}_{\text{pair}}$ contains $961$ images from $613$ subjects ($28.5\%$ of images, $35.1\%$ of subjects; mean GA $176.7 \pm 39.7$ days; TT/TC/TV: $865$/$57$/$39$), and is used exclusively to compute the \rev{alignment module} $\mathcal{F}$.
$\mathcal{D}^{\text{US}}_{\text{train}}$ comprises $1{,}934$ images from $905$ subjects ($57.4\%$ of images, $51.9\%$ of subjects; mean GA $157.0 \pm 45.7$ days; TT/TC/TV: $1{,}514$/$239$/$181$), used to fit the downstream task probes $\theta_{\text{GA}}$ and $\theta_{\text{plane}}$.
$\mathcal{D}^{\text{US}}_{\text{test}}$ contains $474$ images from $226$ subjects ($14.1\%$ of images, $13.0\%$ of subjects; mean GA $161.8 \pm 51.1$ days; TT/TC/TV: $377$/$57$/$40$), and serves as the held-out evaluation set.

$\mathcal{D}^{\text{MRI}}$ comprises $687$ images from $229$ unique subjects, spanning a gestational age range of $147$--$266$ days ($\approx$21.0--38.0 gestational weeks), with a mean of $214.1 \pm 30.6$ days and a median of $217$ days.
Notably, the dataset is perfectly balanced across the three anatomical planes, with exactly $229$ images per plane (TT, TC, TV), since one 2D slice per standard diagnostic plane is extracted from each subject's SRR volume via atlas-based registration (see Appendix \ref{sec:supp_mri_extraction}).
The MRI cohort covers a more restricted and older gestational age range compared to $\mathcal{D}^{\text{US}}$, consistent with the clinical context in which fetal MRI is typically acquired in the second and third trimesters.
$\mathcal{D}^{\text{MRI}}$ is partitioned by first assigning $\approx$70\% of subjects to $\mathcal{D}^{\text{MRI}}_{\text{pair}}$, then splitting the remaining 30\% 80/20 between $\mathcal{D}^{\text{MRI}}_{\text{train}}$ and $\mathcal{D}^{\text{MRI}}_{\text{test}}$. As with the US cohort, each source dataset contributes proportionally to all three splits.
$\mathcal{D}^{\text{MRI}}_{\text{pair}}$ contains $480$ images from $160$ subjects ($69.9\%$ of images and subjects; mean GA $216.1 \pm 29.7$ days; $160$ images per plane), and is used to derive the \rev{alignment module} $\mathcal{F}$.
$\mathcal{D}^{\text{MRI}}_{\text{train}}$ comprises $165$ images from $55$ subjects ($24.0\%$ of images and subjects; mean GA $212.3 \pm 31.2$ days; $55$ images per plane), used to train the downstream probes.
$\mathcal{D}^{\text{MRI}}_{\text{test}}$ contains $42$ images from $14$ subjects ($6.1\%$ of images and subjects; mean GA $198.9 \pm 35.3$ days; $14$ images per plane), reserved for final evaluation.

\subsection{MRI Slice Extraction}
\label{sec:supp_mri_extraction}

Extracting anatomically consistent 2D slices from 3D fetal MRI volumes requires a principled reference frame, since the standard diagnostic planes (TT, TC, TV) are defined anatomically rather than by a fixed voxel axis.
To this end, each subject's SRR volume is registered to its age-matched template from the Computational Radiology Laboratory (CRL) spatiotemporal fetal brain atlas~\cite{gholipour2017normative}, which provides normative 3D references spanning the gestational age range covered by $\mathcal{D}^{\text{MRI}}$.
Registration is performed as a rigid-body alignment maximizing normalized mutual information between the subject volume and the corresponding atlas template, yielding a transformation that brings each subject into a common anatomical coordinate system. 

The three canonical planes are then identified directly in atlas space and back-projected to the subject volume, from which the corresponding 2D slices are sampled.
All extracted slices were subsequently reviewed with the support of an expert radiographer to verify anatomical consistency and manually fix and tune the extraction to comply to the ISUOG guidelines~\cite{malinger2020isuog}. 

\subsection{Preprocessing and Embedding Extraction}

All samples in $\mathcal{D}$ undergo a standardized preprocessing pipeline to ensure cross-modal spatial consistency.
Each image is first resampled to the minimum isotropic resolution of its modality: $0.05\,\text{mm/px}$ for ultrasound and $0.8\,\text{mm/px}$ for MRI.
After resampling, the foreground region is cropped and resized to the input resolution expected by each backbone, preserving the original aspect ratio and padding with zeros where necessary. This aspect-ratio-preserving resize ensures that identical anatomical structures subtend the same apparent size in the backbone's patch grid regardless of acquisition parameters.
For each backbone, image-level feature vectors are obtained by average-pooling over all patch tokens produced by the Vision Transformer encoder. Formally, given a sequence of $N_p$ patch tokens $\{t_k\}_{k=1}^{N_p}$:
\begin{equation}
    z = \frac{1}{N_p} \sum_{k=1}^{N_p} t_k \in \mathbb{R}^d,
\end{equation}
where $d$ is the hidden dimension of the backbone. The resulting vectors are used directly as input to all alignment and downstream probing experiments, with no further normalization.

\subsection{Cross-Modal Dataset Summary}

Table~\ref{tab:dataset_summary} summarises the complete partition statistics for the combined dataset $\mathcal{D} = \mathcal{D}^{\text{US}} \cup \mathcal{D}^{\text{MRI}}$. In total, $\mathcal{D}$ contains $4{,}056$ images from $1{,}973$ unique subjects across both modalities ($N_{\text{US}}=3{,}369$ from $1{,}744$ subjects; $N_{\text{MRI}}=687$ from $229$ subjects).

\begin{table}
\centering
\caption{Dataset partition statistics for $\mathcal{D}^{\text{US}}$ and $\mathcal{D}^{\text{MRI}}$. GA is reported in days as mean $\pm$ std. \% columns refer to the fraction of the respective modality total. Plane counts are TT / TC / TV.}
\label{tab:dataset_summary}
\resizebox{\textwidth}{!}{%
\begin{tabular}{llcccccc}
\toprule
\textbf{Modality} & \textbf{Split} & \textbf{\#Images} & \textbf{\%~Img} & \textbf{\#Subjects} & \textbf{\%~Subj} & \textbf{GA (days)} & \textbf{TT / TC / TV} \\
\midrule
\multirow{4}{*}{US}
 & Overall                       & 3{,}369 & ---   & 1{,}744 & ---   & $163.3 \pm 45.7$ & 2756 / 353 / 260 \\
 & $\mathcal{D}_{\text{pair}}$  & 961     & 28.5  & 613     & 35.1  & $176.7 \pm 39.7$ & 865 / 57 / 39    \\
 & $\mathcal{D}_{\text{train}}$ & 1{,}934 & 57.4  & 905     & 51.9  & $157.0 \pm 45.7$ & 1514 / 239 / 181 \\
 & $\mathcal{D}_{\text{test}}$  & 474     & 14.1  & 226     & 13.0  & $161.8 \pm 51.1$ & 377 / 57 / 40    \\
\midrule
\multirow{4}{*}{MRI}
 & Overall                       & 687     & ---   & 229     & ---   & $214.1 \pm 30.6$ & 229 / 229 / 229  \\
 & $\mathcal{D}_{\text{pair}}$  & 480     & 69.9  & 160     & 69.9  & $216.1 \pm 29.7$ & 160 / 160 / 160  \\
 & $\mathcal{D}_{\text{train}}$ & 165     & 24.0  & 55      & 24.0  & $212.3 \pm 31.2$ & 55 / 55 / 55     \\
 & $\mathcal{D}_{\text{test}}$  & 42      & 6.1   & 14      & 6.1   & $198.9 \pm 35.3$ & 14 / 14 / 14     \\
\bottomrule
\end{tabular}%
}
\end{table}

\section{Implementation Details}
\label{sec:supp_impl}

\subsection{Entropic Optimal Transport}

The transport plan \rev{$P$} is computed via a log-domain Sinkhorn--Knopp algorithm run for $100$ iterations with uniform marginals \rev{$\mathbf{a} = \frac{1}{N_s}\mathbf{1}_{N_s}$ and $\mathbf{b} = \frac{1}{N_t}\mathbf{1}_{N_t}$, where $N_s$ and $N_t$} are the number of source and target samples in $\mathcal{D}_{\text{pair}}$, respectively.
The implementation uses GPU acceleration.
Before computing the soft target anchors \rev{$\tilde{X}^{\mathcal{T}} = \bar{P} X^{\mathcal{T}}$}, each row of ${P}$ is normalized to sum to one (giving \rev{$\bar{P}$}), so that each source sample receives a convex combination of target embeddings.

\minisection{Cost Matrix Construction.}
The composite cost matrix $\tilde{C} = C - \min(C)$  is assembled from three terms. The temporal term \rev{$D^{\text{time}}_{i,j}$} $= |g^{\mathcal{S}}_i - g^{\mathcal{T}}_j|$ measures the absolute gestational age discrepancy in days. The embedding term $\rev{D^{\text{proto}}_{i,j}}$ measures the Euclidean distance between the plane-and-age-binned centroids of $X^{\mathcal{T}}$ associated with sample $i$ and sample $j$: target embeddings are grouped into bins of width \rev{$\Delta$} days within each anatomical plane, per-bin centroids are computed, and each sample is mapped to the centroid of the nearest available bin within its plane.
The spatial term \rev{$D^{\text{space}}_{i,j}$} $\in \{0, 1\}$ is a binary indicator that equals $1$ whenever source sample $i$ and target sample $j$ belong to different anatomical planes.
The two continuous terms are independently z-scored before combining:
\begin{equation}
    \rev{C = \mathcal{Z}(D^{\text{time}}) + \mathcal{Z}(D^{\text{proto}}) + D^{\text{space}},}
\end{equation}
\rev{where $\mathcal{Z}(\cdot)$ standardizes the entries of a matrix to zero mean and unit variance, as in Eq.~\eqref{eq:cost}.} The result is shifted by its minimum value to ensure non-negativity before being passed to the Sinkhorn solver.

\minisection{Hyperparameter Search Space.}
The entropy regularization $\varepsilon$ is searched over $\{10^{-4}, 5\times10^{-4}, 10^{-3}, 5\times10^{-3}\}$ and the centroid bin width \rev{$\Delta$} over $\{7, 14, 21\}$ days. All cost term weights are fixed to $1.0$ throughout.

\subsection{Alignment Models}
In the main paper we evaluate the following alignment methods:
\begin{itemize}
    \item \textbf{Ridge.}
The regularization coefficient $\lambda$ is searched over $\{10^{-4}, 5\times10^{-4}, 10^{-3}, 5\times10^{-3}, 10^{-2}, 5\times10^{-2}, 10^{-1}, 5\times10^{-1}\}$.
    \item \textbf{CCA.}
The full-rank solution is used (no dimensionality reduction). Both within-modality covariance matrices are regularized by adding \rev{$\eta \mathbf{I}$ with $\eta = 10^{-6}$} before inversion.
    \item \textbf{CORAL.}
The covariance matrices are regularized with \rev{$\eta = 10^{-5}$}. \rev{CORAL is evaluated as a standalone pairing-free alignment baseline.}
    \item \textbf{MLP.}
The hidden layer dimension is searched over $\{256, 512, 1024\}$. When source and target embedding dimensions coincide, a residual connection is added around the hidden layer. Training uses learning rate $10^{-3}$ and weight decay searched over $\{10^{-4}, 5\times10^{-4}, 10^{-3}, 5\times10^{-3}, 10^{-2}, 5\times10^{-2}, 10^{-1}, 5\times10^{-1}\}$, with a maximum of $600$ epochs and early stopping with patience $80$ (epochs without improvement on the training MSE). Weights are initialized with Kaiming uniform initialization. Training is performed on a single NVIDIA A100 GPU.

\end{itemize}

\subsection{Downstream Probes}

The GA Ridge regressor is fit with regularization coefficient $\alpha = 1.0$. The plane SVM uses balanced class weights to compensate for the plane imbalance in $\mathcal{D}^{\text{US}}$ described in Sec. \ref{sec:supp_data}.

\FloatBarrier
\section{Additional Experiments}
\label{sec:supp_addl}
\label{sec:supp_rebuttal}
\rev{All results in this section are on the held-out test sets described in Appendix~\ref{sec:supp_data}. CORAL, CCA and Ridge are deterministic given the pairing; unless stated otherwise, MLP rows report mean $\pm$ std over 3 seeds, and CCA and Ridge rows with Random pairing are single runs. Sec.~\ref{sec:supp_mlp_hyper} studies the sensitivity of MLP alignment to its hyperparameters, Sec.~\ref{sec:supp_mricore} reports the MRI-native MRI-CORE in both directions and as a distillation student, and Secs.~\ref{sec:supp_unpaired} and~\ref{sec:supp_finetune} compare \method with unpaired alignment baselines and with direct fine-tuning.}

\subsection{\rev{MLP Alignment Hyperparameters}}
\label{sec:supp_mlp_hyper}
Tab.~\ref{tab:mlp_hyperparam_ablation} reports the sensitivity of MLP alignment to the four hyperparameters of \method: the entropy regularization $\varepsilon$, the centroid bin width \rev{$\Delta$}, the \rev{weight decay}, and the hidden dimension of the network.
The two dominant factors are \rev{$\Delta$} and $\varepsilon$, which interact strongly. Larger bin widths produce smoother target-space prototypes that better represent the local geometry of the embedding manifold; combined with a sufficiently large $\varepsilon$ ($5\times10^{-3}$), which yields moderately diffuse transport and thus more stable soft anchors, this drives the largest improvements in GA regression (\rev{$\Delta=21$}, $\varepsilon=5\times10^{-3}$: MAE $18.3$ days). 
In contrast, the \rev{weight decay} and the hidden dimension show negligible sensitivity: varying the \rev{weight decay} across its entire search range changes mean MAE by at most $0.3$ days and mean Cls.\ by $0.5\%$, while varying the hidden dimension yields differences below $0.2$ days and $1.1\%$ respectively.
Across all $288$ configurations, $96\%$ improve over the Zero-Shot MAE baseline and $83\%$ improve over the Zero-Shot plane classification accuracy, confirming that MLP alignment is robust to hyperparameter choice.
\begin{table}[!ht]
\centering
\caption{MLP alignment hyperparameter sensitivity (USFM, MRI$\to$US, test set).
Each cell reports GA Reg.\,$\downarrow$ (days) / Plane Cls.\,$\uparrow$ (\%), averaged over
\rev{weight decay} $\in \{10^{-4},\ldots,5\!\times\!10^{-1}\}$ and hidden dim $\in \{256,512,1024\}$.
\textbf{Bold} = column-wise best.}
\label{tab:mlp_hyperparam_ablation}
\small
\setlength{\tabcolsep}{6pt}
\renewcommand{\arraystretch}{1.08}
\begin{tabular}{l cc cc cc cc}
\toprule
& \multicolumn{2}{c}{$\varepsilon = 10^{-4}$}
& \multicolumn{2}{c}{$\varepsilon = 5\!\times\!10^{-4}$}
& \multicolumn{2}{c}{$\varepsilon = 10^{-3}$}
& \multicolumn{2}{c}{$\varepsilon = 5\!\times\!10^{-3}$} \\
\cmidrule(lr){2-3}\cmidrule(lr){4-5}\cmidrule(lr){6-7}\cmidrule(lr){8-9}
\rev{$\Delta$} (days)
  & GA\,$\downarrow$ & Cls.\,$\uparrow$
  & GA\,$\downarrow$ & Cls.\,$\uparrow$
  & GA\,$\downarrow$ & Cls.\,$\uparrow$
  & GA\,$\downarrow$ & Cls.\,$\uparrow$ \\
\midrule
$7$  & 21.8 & 64.4 & 21.4 & \textbf{66.0} & 21.4 & \textbf{66.5} & 20.7 & 66.6 \\
$14$ & \textbf{20.4} & 64.3 & 20.2 & 60.0 & 20.4 & 59.1 & 20.9 & 67.7 \\
$21$ & 20.8 & \textbf{65.7} & \textbf{20.0} & 65.6 & \textbf{19.8} & 65.3 & \textbf{18.3} & \textbf{68.3} \\
\bottomrule
\end{tabular}
\end{table}

\FloatBarrier
\subsection{\rev{MRI-CORE}}
\label{sec:supp_mricore}

\minisection{\rev{Pairing ablation in both directions.}}
\rev{Tab.~\ref{tab:supp_mricore} reports MRI-CORE in both directions. In US$\to$MRI, US is the source and MRI the native target space (the MRI-CORE column of Tab.~\ref{tab:self_distill_us2mri}); in MRI$\to$US, the source is MRI-CORE's native modality, mapped toward the US space. In both directions, Random and Cosine pairing underperform \method across CCA, Ridge, and MLP alignment, with the gap being particularly large for GA regression. In US$\to$MRI, \method+MLP gives the best plane accuracy ($98.2\pm0.4\%$) with a GA MAE of $12.8\pm0.2$ days, close to Zero-Shot ($12.3$). In MRI$\to$US, Zero-Shot is already strong on GA ($6.5$ days) and no alignment improves it; \method+MLP stays closest ($7.0\pm1.0$ days), and \method+CCA gives the best plane accuracy ($81.0\%$ vs.\ $76.2\%$ Zero-Shot).}

\minisection{\rev{Teacher and transfer-head ablation.}}
\rev{Tab.~\ref{tab:supp_mricore_distill} ablates both the teacher representation (Raw vs.\ +\method) and the transfer head (Ridge vs.\ MLP) when distilling into MRI-CORE. Across both SAM-Med2d and BioMedParse teachers, MLP outperforms Ridge on GA regression. Applying \method to the teacher is beneficial only with the MLP head: for SAM-Med2d, GA improves from $12.3$ to $11.6$ days, whereas the corresponding Ridge variant degrades from $13.3$ to $14.5$ days. A similar interaction appears for BioMedParse, where +\method with MLP preserves the $12.0$-day GA result while +\method with Ridge degrades to $15.6$ days. This ablation shows that the effect of teacher refinement depends strongly on the transfer-head choice.}

\begin{table}[!ht]
\centering
\small
\caption{\rev{MRI-CORE in both directions. US$\to$MRI is evaluated on the US test set, MRI$\to$US on the MRI test set. Best and second-best results in each metric column are shown in \textbf{bold} and \underline{underline}, respectively.}}
\label{tab:supp_mricore}
\setlength{\tabcolsep}{4pt}
\begin{tabular}{l l cc cc}
\toprule
& & \multicolumn{2}{c}{\rev{US$\to$MRI}} & \multicolumn{2}{c}{\rev{MRI$\to$US}} \\
\cmidrule(lr){3-4} \cmidrule(lr){5-6}
Pairing & Alignment & GA\,$\downarrow$ & Cls.\,$\uparrow$ & GA\,$\downarrow$ & Cls.\,$\uparrow$ \\
\midrule
Zero-Shot & --- & \textbf{12.3} & \underline{98.1} & \rev{\textbf{6.5}} & \rev{76.2} \\
--- & CORAL & 16.7 & 96.2 & \rev{19.8} & \rev{57.1} \\
\midrule
\multirow{3}{*}{Greedy} & CCA & 24.9 & 93.5 & \rev{13.6} & \rev{73.8} \\
 & Ridge & 25.3 & 93.5 & \rev{9.1} & \rev{73.8} \\
 & MLP & \textbf{12.3$\pm$0.2} & 97.2$\pm$0.1 & \rev{9.0$\pm$0.1} & \rev{78.6$\pm$2.4} \\
\midrule
\multirow{3}{*}{Hungarian} & CCA & 16.1 & 92.0 & \rev{16.5} & \rev{71.4} \\
 & Ridge & 14.2 & 96.8 & \rev{15.8} & \rev{69.0} \\
 & MLP & 13.7$\pm$0.2 & 97.4$\pm$0.1 & \rev{8.9$\pm$0.7} & \rev{\underline{79.4$\pm$1.4}} \\
\midrule
\multirow{3}{*}{Random} & CCA & 27.6 & 81.9 & \rev{29.5} & \rev{42.9} \\
 & Ridge & 27.7 & 81.2 & \rev{24.5} & \rev{47.6} \\
 & MLP & 20.2$\pm$0.7 & 95.7$\pm$0.6 & \rev{21.6$\pm$0.6} & \rev{55.6$\pm$5.5} \\
\midrule
\multirow{3}{*}{Cosine} & CCA & 30.1 & 75.3 & \rev{25.4} & \rev{57.1} \\
 & Ridge & 30.9 & 74.7 & \rev{26.6} & \rev{54.8} \\
 & MLP & 20.7$\pm$0.7 & 95.9$\pm$0.8 & \rev{20.8$\pm$1.0} & \rev{73.0$\pm$3.6} \\
\midrule
\rowcolor{blue!8} & CCA & 24.6 & 92.6 & \rev{15.2} & \rev{\textbf{81.0}} \\
\rowcolor{blue!8} \method & Ridge & 25.1 & 93.0 & \rev{13.4} & \rev{76.2} \\
\rowcolor{blue!8} & MLP & \underline{12.8$\pm$0.2} & \textbf{98.2$\pm$0.4} & \rev{\underline{7.0$\pm$1.0}} & \rev{77.0$\pm$1.4} \\
\bottomrule
\end{tabular}
\end{table}

\begin{table}[!ht]
\centering
\small
\caption{\rev{MRI-CORE as student, US$\to$MRI. Teachers are improved with \method in the US$\to$MRI direction, and $\mathcal{G}_{\text{US}\to\text{US}}$ (Ridge or MLP) maps MRI-CORE's US embeddings to the teacher's; evaluation is on the US test set. Raw rows are deterministic; + \method rows, including Ridge ones, report mean $\pm$ std over the 3 seeds of the teacher alignment. Best and second-best results in each metric column are shown in \textbf{bold} and \underline{underline}, respectively.}}
\label{tab:supp_mricore_distill}
\setlength{\tabcolsep}{4pt}
\begin{tabular}{l l l c c}
\toprule
Teacher & Setup & $\mathcal{G}$ & GA\,$\downarrow$ & Cls.\,$\uparrow$ \\
\midrule
Zero-Shot & --- & --- & 12.3 & \textbf{98.1} \\
\midrule
\multirow{4}{*}{SAM-Med2d} & Raw & Ridge & 13.3 & 97.3 \\
 & Raw & MLP & 12.3$\pm$0.1 & 97.5$\pm$0.2 \\
 & + \method & Ridge & 14.5$\pm$0.4 & 95.2$\pm$0.7 \\
 & + \method & MLP & \textbf{11.6$\pm$0.1} & \underline{98.0$\pm$0.1} \\
\midrule
\multirow{4}{*}{BioMedParse} & Raw & Ridge & 13.2 & 97.9 \\
 & Raw & MLP & \underline{12.0$\pm$0.3} & 97.6$\pm$0.2 \\
 & + \method & Ridge & 15.6$\pm$0.2 & 96.5$\pm$0.1 \\
 & + \method & MLP & \underline{12.0$\pm$0.1} & 97.7$\pm$0.4 \\
\bottomrule
\end{tabular}
\end{table}

\FloatBarrier
\begingroup
\subsection{Unpaired Alignment Baselines}
\label{sec:supp_unpaired}
We compare against two representative unpaired approaches under the same frozen-backbone protocol (Tab.~\ref{tab:supp_unpaired}). \textit{MRI$\to$US synthesis}: a multimodal hierarchical VAE~\cite{dorent2026unified} translates fetal MRI slices into synthetic US images, which go through the same preprocessing as all other images (Appendix~\ref{sec:supp_data}) and are embedded by each frozen backbone; the probes are fit on the synthetic-US training split and evaluated on the same 14 held-out MRI test subjects. \textit{Adversarial alignment}: a feature-level variant of ADDA~\cite{tzeng2017adversarial}, where a trainable mapper pushes MRI embeddings toward independently sampled US embeddings to fool a discriminator (two 500-unit ReLU layers, batch size 64, 600 alternating updates, AdamW, multiple seeds), without metadata or subject pairing. Training did not collapse, but domain confusion was never reached: the discriminator stayed 73--98\% accurate, and a held-out classifier separated transformed MRI from real US embeddings with $>$99\% accuracy. The mapper is not the identity, yet the downstream metrics stay at Zero-Shot level. The synthesis baseline improves USFM plane accuracy but degrades Echocare, SAM-Med2d and BioMedParse on both tasks.

\begin{table}[!ht]
\centering
\caption{\rev{\textbf{MRI$\to$US.}} Unpaired alignment baselines on the MRI test set. \method results are in Tab.~\ref{tab:self_distill_mri2us}. Best and second-best results within each backbone and metric are shown in \textbf{bold} and \underline{underline}, respectively.}
\label{tab:supp_unpaired}
\small
\setlength{\tabcolsep}{4pt}
\begin{tabular}{l cc cc cc}
\toprule
& \multicolumn{2}{c}{Zero-Shot} & \multicolumn{2}{c}{MRI$\to$US synthesis} & \multicolumn{2}{c}{Adversarial alignment} \\
\cmidrule(lr){2-3} \cmidrule(lr){4-5} \cmidrule(lr){6-7}
Backbone & GA\,$\downarrow$ & Cls.\,$\uparrow$ & GA\,$\downarrow$ & Cls.\,$\uparrow$ & GA\,$\downarrow$ & Cls.\,$\uparrow$ \\
\midrule
USFM
& 21.9 & 61.9
& \textbf{20.8} & \textbf{71.4}
& \underline{21.8} & \underline{64.3} \\
Echocare
& \textbf{10.3} & \textbf{81.0}
& \underline{10.7} & \underline{71.4}
& \textbf{10.3} & \textbf{81.0} \\
SAM-Med2d
& \textbf{7.7} & \textbf{83.3}
& \underline{9.5} & \underline{81.0}
& \textbf{7.7} & \textbf{83.3} \\
BioMedParse
& \textbf{6.5} & \textbf{85.7}
& 8.6 & \underline{81.0}
& \underline{6.6} & \textbf{85.7} \\
\bottomrule
\end{tabular}
\end{table}

\FloatBarrier
\subsection{Fine-Tuning on MRI}
\label{sec:supp_finetune}
As a reference outside our setting, we fine-tune each backbone directly on the MRI training split (Tab.~\ref{tab:supp_finetune}, mean $\pm$ std over 3 seeds). Fine-tuning lowers the MRI GA error further than \method, at the price of modifying the backbone: it increases the GA error on the original US domain for every backbone and reduces Echocare's US plane accuracy. \method keeps the backbone frozen and preserves the source-domain representation by construction.

\begin{table}[!ht]
\centering
\caption{Direct fine-tuning on MRI. Left: MRI test set. Right: US test set before and after fine-tuning. Best and second-best results within each backbone, test domain, and metric are shown in \textbf{bold} and \underline{underline}, respectively.}
\label{tab:supp_finetune}
\small
\setlength{\tabcolsep}{4pt}
\begin{tabular}{l cc cc cc cc}
\toprule
& \multicolumn{4}{c}{MRI test set} & \multicolumn{4}{c}{US test set} \\
\cmidrule(lr){2-5} \cmidrule(lr){6-9}
& \multicolumn{2}{c}{Zero-Shot} & \multicolumn{2}{c}{Fine-tuned} & \multicolumn{2}{c}{Before} & \multicolumn{2}{c}{After fine-tuning} \\
\cmidrule(lr){2-3} \cmidrule(lr){4-5} \cmidrule(lr){6-7} \cmidrule(lr){8-9}
Backbone & GA\,$\downarrow$ & Cls.\,$\uparrow$ & GA\,$\downarrow$ & Cls.\,$\uparrow$ & GA\,$\downarrow$ & Cls.\,$\uparrow$ & GA\,$\downarrow$ & Cls.\,$\uparrow$ \\
\midrule
USFM
& \underline{21.9} & \underline{61.9}
& \textbf{8.4$\pm$1.4} & \textbf{81.7$\pm$11.3}
& \textbf{16.0} & \underline{92.8}
& \underline{18.3$\pm$0.3} & \textbf{95.1$\pm$1.2} \\
Echocare
& \underline{10.3} & \underline{81.0}
& \textbf{7.9$\pm$2.0} & \textbf{86.5$\pm$3.6}
& \textbf{14.9} & \textbf{95.6}
& \underline{26.0$\pm$3.0} & \underline{88.2$\pm$4.2} \\
SAM-Med2d
& \underline{7.7} & \underline{83.3}
& \textbf{5.3$\pm$0.5} & \textbf{88.1$\pm$2.4}
& \textbf{15.7} & \underline{96.2}
& \underline{16.8$\pm$0.5} & \textbf{98.6$\pm$0.5} \\
BioMedParse
& \underline{6.5} & \underline{85.7}
& \textbf{4.3$\pm$0.2} & \textbf{92.1$\pm$1.4}
& \textbf{12.1} & \underline{98.7}
& \underline{15.0$\pm$1.5} & \textbf{98.9$\pm$0.9} \\
\bottomrule
\end{tabular}
\end{table}
\endgroup
\FloatBarrier

\clearpage

\section{Fetal Brain Development}
\label{sec:appendix_biology}
The Weakly Aligned Spatiotemporal Pairs (WASP) framework is fundamentally anchored in the predictable, stereotyped trajectory of fetal brain development. In neurotypical gestations, the structural maturation of the fetal brain produces highly consistent macroscopic morphologies at any given gestational age (GA) \cite{gholipour2017normative}. This biological rigidity across subjects allows WASP to establish valid cross-modal correspondences using only time (GA) and space (diagnostic plane) as metadata.
\subsection{The Spatiotemporal Trajectory of Neural Development}
\begin{wrapfigure}[26]{r}{.5\textwidth}
    \centering
    \vspace{-.5cm}    
    \includegraphics[width=0.4\textwidth]{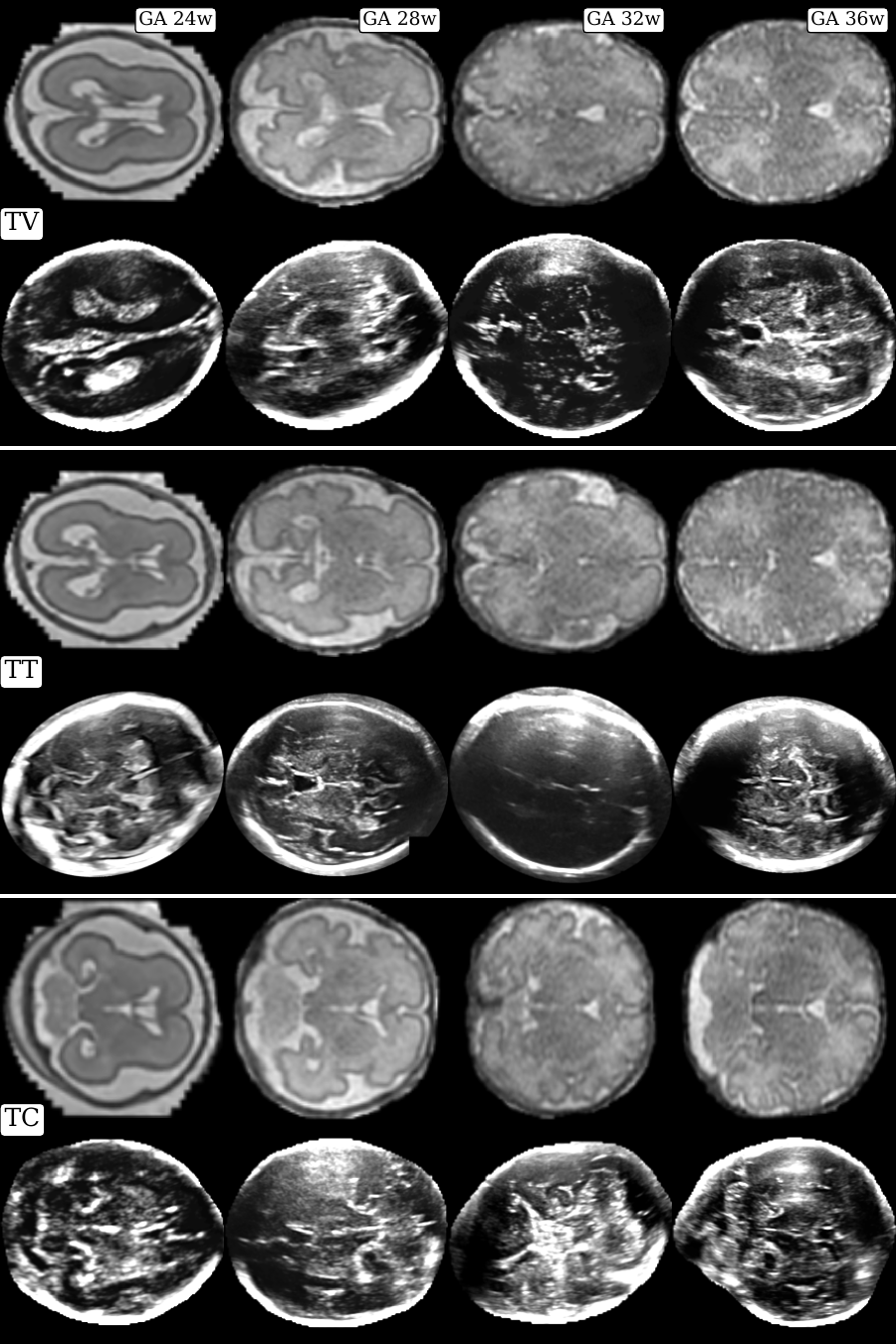}
    \caption{Spatiotemporal Evolution of Fetal Brain Anatomy in MRI and US. Rows represent the three standard diagnostic planes (TV, TT, TC). Columns represent different Gestational Ages.}
    \label{fig:spatiotemporal_biology}
\end{wrapfigure}
Fetal brain maturation during the second and third trimesters is characterized by explosive volumetric growth, neuronal migration, and the initiation of complex sulcation and gyration \cite{govaert2020developing,moltoni2021brain}. Early in the second trimester (approx. 14--16 weeks GA), the telencephalic vesicles are relatively smooth (lissencephalic), and the cortical plate is just beginning to organize via the migration of neuroblasts along radial glia \cite{rakic1972mode,subramanian2020cortical}. 
As gestation progresses into the 20--28 week window—the primary overlap period for both the US and MRI cohorts in this study—the brain transitions from a smooth state to a highly convoluted structure. The primary sulci (such as the lateral/Sylvian fissure, parieto-occipital sulcus, and central sulcus) appear in a strict temporal sequence \cite{chi1977gyral,garel2001fetal}. 
Because the timing of these invaginations is genetically and developmentally constrained, the sulcal pattern serves as an accurate biological clock. Consequently, the anatomical variance between two neurotypical subjects at the exact same GA is lower than the variance within a single subject observed just weeks apart \cite{dubois2019dynamics,dubois2008primary}. This constraint is what minimizes the $D^{\text{time}}$ and $\rev{D^{\text{proto}}}$ costs in the WASP Optimal Transport formulation.

\subsection{Anatomical Grounding of the Diagnostic Planes}
Standardized diagnostic planes capture distinct axial cross-sections of this developing 3D anatomy. The spatial cost term (\rev{$D^{\text{space}}$}) in WASP enforces strict plane matching because each plane intercepts entirely different neurobiological distinct structures \cite{malinger2020isuog}:
\begin{itemize}
    \item \textbf{Transventricular Plane (TV):} Intercepts the superior aspect of the lateral ventricles. Early in gestation, the choroid plexus is highly prominent and fills the lateral ventricles. As the cortical mantle thickens due to arborization and synaptogenesis, the relative volume of the lateral ventricles decreases \cite{hu2025choroid, malinger2020isuog}.
    \item \textbf{Transthalamic Plane (TT):} A slightly lower axial cut that captures the thalami, the midline falx, and the cavum septi pellucidi (CSP). The CSP is a critical fluid-filled space between the leaflets of the septum pellucidum; its typical development and subsequent obliteration later in life are key markers of normal midline forebrain fusion \cite{winter2010cavum, nagaraj2018abnormalities, malinger2020isuog}.
    \item \textbf{Transcerebellar Plane (TC):} Angled posteriorly to capture the posterior fossa, including the cerebellar hemispheres, the vermis, and the cisterna magna. The cerebellum undergoes dramatic foliation and exponential growth in the late second and third trimesters, which is highly visible in both US and MRI \cite{scott20123d, liu2011development}.
\end{itemize}

\section{Broader Impact}
\label{sec:supp_broader_impact}
Although our study focuses on the alignment of fetal MRI and ultrasound, the core principles of \method extend far beyond this specific use case. By demonstrating that routine clinical metadata can successfully substitute for subject-level paired scans, our framework provides a versatile blueprint for cross-modal representation learning across diverse medical imaging domains.

Concrete candidates for the same construction include pediatric MRI/CT/US studies indexed by chronological age and acquisition protocol, adult MRI/PET cohorts indexed by anatomical landmarks, and histopathology--radiology pairings indexed by tumor stage. In each case the alignment cost is built from coordinates that are already part of routine clinical reporting, so no new annotation effort is needed beyond what hospitals already record.

Two caveats specific to these broader uses are worth stating. First, \method establishes anatomical, not subject-level, correspondence: matched scans share a developmental coordinate but not an identity, and must not be interpreted as same-subject evidence in an individual workup. Second, because the underlying backbones are reused as-is, biases present in their pretraining cohorts (scanner vendor, geography, age range) propagate into the aligned representations and should be audited rather than assumed to be neutralized by the alignment step.

\end{appendices}

\end{document}